\documentclass[11pt]{article}
\usepackage{subcaption}
\usepackage{amsfonts}
\usepackage{amsmath}

\usepackage[preprint]{acl}

\usepackage{times}
\usepackage{latexsym}
\usepackage{booktabs}
\usepackage{multirow}
\usepackage{array}
\usepackage{mathtools}

\usepackage[T1]{fontenc}

\usepackage[utf8]{inputenc}

\usepackage{microtype}

\usepackage{inconsolata}

\usepackage{graphicx}
\graphicspath{{latex/}{./}}

\usepackage{enumerate}
\usepackage{enumitem}

\title{Locating and Controlling Implicit \\Personalization in Large Language Models}

\author{
Yueru Yan \quad Siqi Wu \quad Thai Le \\
Indiana University \\
Bloomington, USA \\
\texttt{\{yueryan,swu2,tle\}@iu.edu}
}

\begin{document}
\maketitle
\begin{abstract}
Large language models (LLMs) often shift their outputs in response to implicit demographic cues even when users never state a demographic identity. Previous work has documented this behavior, but the connection between these behavioral changes and the model's internal activations remains unclear. Using matched cued and neutral conversations across five LLMs, we establish that a localized internal activation signal tracks changes in recommendations, with correlations up to $r=0.87$. When multiple cues appear together, their internal signals largely combine, but the changes in output do not simply add up. We further show that removing the internal signal associated with one cue can suppress its influence, often more effectively than asking the model to ignore demographics via prompting, while largely preserving general benchmark performance. However, the ability to selectively remove one dimension's influence while leaving co-present dimensions intact remains highly model- and attribute-specific. These results connect implicit personalization behavior to an internal signal that can be analyzed and causally controlled.

\end{abstract}

\section{Introduction}

\label{sec:intro}

When a user mentions a Kwanzaa celebration, slips in \emph{``no cap''},
or asks about playground games, a large language model (LLM) may quietly
tailor its subsequent recommendations in ways aligned with demographic
associations carried by those cues, a
phenomenon called \emph{implicit personalization}
\citep{jin2024implicit, kantharuban2025stereotype}. Previous studies have
demonstrated this phenomenon across model families: identity cues woven
into multi-turn conversation reliably move recommendations toward
stereotype-aligned content \citep{kotek2023gender, hofmann2024ai,
neplenbroek2025reading, gupta2024bias, etgar2024implicit}. The resulting
harms are concrete and consequential. Dialect features alone have been
shown to elicit harsher hypothetical criminal sentences and
lower-prestige job assignments \citep{hofmann2024ai}; recommendations
narrow toward stereotypical content when implicit demographic correlates
appear in user history \citep{kantharuban2025stereotype,
chen2024designing}. As LLMs are increasingly integrated into critical decision-making pipelines, this silent adaptation becomes especially problematic, leaving users with no mechanism to see the conditioning signal, contest it, or opt out of it.

\begin{figure*}[t]
\centering
\includegraphics[width=0.97\linewidth]{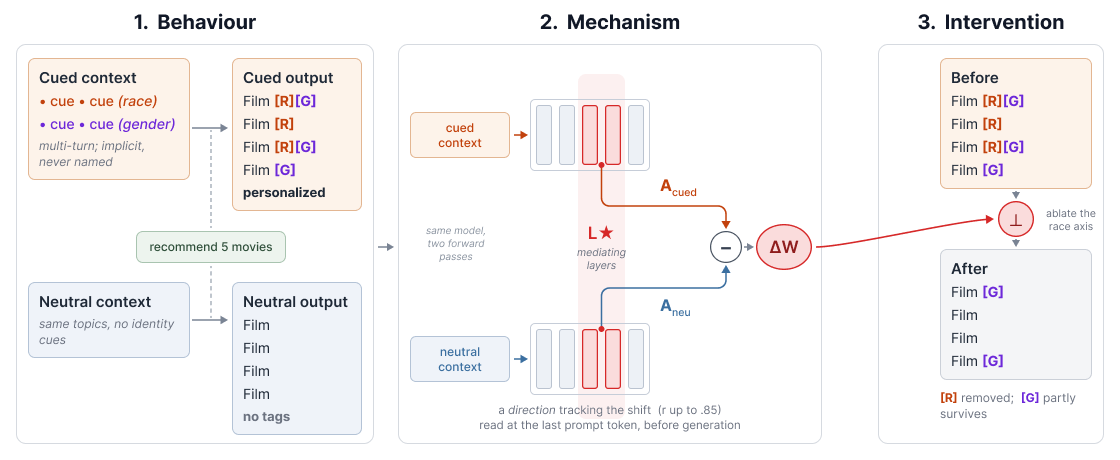}
\caption{Overview. Implicit cues shift recommendations, and the
cue-induced activation contrast at $L^\star$ (identified mediated layers) tracks the shift. Furthermore, projecting its direction out suppresses it.}
\label{fig:overview}
\end{figure*}

While the behavioral consequences of implicit personalization are well-established~\cite{neplenbroek2025reading}, it remains largely unclear how this shift relates to the model's internal representations. Some studies touch the intermediate activations by adopting probe-based methods to predict demographic attributes associated with a user's context \citep{chen2024designing, ghandeharioun2024s}. Such probes show that attribute information is
decodable, but not whether it causally shapes the answer.
Additionally, real conversations also carry several cues at once. A single dialogue
might contain cultural references and age-specific slang together, so a
single-axis account may not predict what a model does when cues overlap.
We therefore ask: does the internal representation of a
mixed-cue context decompose into its single-cue parts, and does the behavior it drives do the same?


We make this connection through the raw cue-induced activation contrast
$\Delta\mathbf A$, the residual-stream difference between matched cued and neutral contexts at the specific layers that mediate the behavioral shift.
Figure~\ref{fig:overview} summarizes the three steps: the cued and neutral
dialogues produce different recommendations, their activation difference at these localized layers tracks how large that difference is, and projecting that
difference out of the residual stream during generation suppresses it.
We evaluate five LLMs (7B to 14B parameters) on movie recommendation,
with $n=50$ matched cued--neutral dialogue pairs for each of nine
single-cue conditions and each of three two-cue combinations of race,
gender, and age.


Our main contributions are: 

\begin{enumerate}[leftmargin=\dimexpr\parindent-0.01\labelwidth\relax, noitemsep, topsep=0pt]
    \item \textbf{An internal signal that tracks personalization.} We find that the normalized magnitude of the activation contrast ($\Delta A$) strongly correlates with the behavioral shift on a per-sample basis (up to $r=.87$) and falls toward zero in conditions where no shift occurs. 
    \item \textbf{Mapping how multiple identities combine.} When several demographic cues appear at once, the mixed-cue contrast $\Delta A$ can be reliably reconstructed from the sample's own single-cue contrasts. However, the final output does not perfectly mirror this internal addition: the actual behavioral shift is sub-additive, falling $28$--$38\%$ below the sum of the single-cue shifts.
    \item \textbf{Turning off implicit personalization in multi-cue
    contexts.} By projecting out the internal direction associated with a specific demographic cue during generation, we can selectively suppress its influence on the output on some models and dimensions. This targeted internal ablation matches or beats explicit ``ignore demographics'' prompting, which occasionally backfires and increases stereotyping, while largely preserving the model's general reasoning capabilities.
\end{enumerate}

Together, these results establish the missing connection between implicit
personalization and the model's internal representation: the normalized magnitude of $\Delta\mathbf A$ at targeted
layers ($L^\star$) tracks the shift a model would produce, and its
sample-averaged direction provides an inference-time target for suppressing
that shift.

\section{Related Work}

Implicit personalization is closely related to demographic bias, but the
shift is driven by cues a user never states rather than by a supplied
attribute. Early benchmarks such as StereoSet \citep{nadeem2021stereoset},
BOLD \citep{dhamala2021bold}, and CEB \citep{wang2025ceb} characterize
stereotype associations in generated text, and more recent audits study how
identity-correlated user information changes task outputs in reference
letters \citep{wan2023kelly}, occupational and resume settings
\citep{gelling2026auditing, van2025race, wilson2024gender}, financial advice
\citep{etgar2024implicit}, and dialect-conditioned decisions
\citep{hofmann2024ai}. The key distinction is whether identity is stated or
only implied: explicit audits supply a demographic persona
\citep{kantharuban2025stereotype, van2025race}, whereas implicit audits vary
identity-correlated cues without naming the identity
\citep{jin2024implicit, neplenbroek2025reading}. Such shifts are plausible
because frontier models can recover personal attributes from ordinary text
\citep{staab2024beyond}. Purely behavioral audits nevertheless have limits:
different cue formats yield different conclusions, and overlapping identities produce effects invisible to single-axis testing \citep{ma2023intersectional,
wilson2024gender}.

Mechanism-level work has examined how user attributes are represented
internally, through linear probes \citep{neplenbroek2025reading} and
persona-tracking systems \citep{ghandeharioun2024s, chen2024designing}.
Probing shows that attribute information is decodable, not that it drives
the output. Following the causal mediation tradition
\citep{vig2020investigating}, we instead ask whether a quantity read from
the cue-induced activation contrast tracks the behavioral shift and supports
targeted intervention.

A complementary literature shows that activation directions can change
behavior directly: representation engineering builds contrast directions to
steer generation \citep{zou2023representation, li2023inference,
rodriguez2025controlling, yu2025understanding, arditi2024refusal}, and
related work erases protected attributes by iterative projection
\citep{ravfogel2020null} or closed-form linear erasure
\citep{belrose2023leace}. Closest to our intervention, neutralizing
attribute directions outperforms anti-bias prompts in explicit hiring
\citep{karvonen2025robustly}, and traits can be edited with capability
preserved \citep{ju2025probing}. These methods build directions from stated
attributes or labeled probes. We take the contrast induced by naturally
occurring implicit cues and ask whether its magnitude tracks the behavioral
shift, whether it composes when cues co-occur, and whether removing its
direction suppresses the behavior, connecting the disparities reported in
fairness audits of LLM recommenders \citep{zhang2023chatgpt,
kantharuban2025stereotype} to a measurable internal quantity.

\section{Methodology}
\label{sec:methodology}


\subsection{Dataset}
\label{sec:dataset}

We use a paired-comparison framework in which each sample consists of matched multi-turn dialogues that differ only in whether they contain implicit demographic cues. We generate all dialogues using GPT-4o~\cite{hurst2024gpt}. To fully capture the personalization effect, we draw cues from a literature-grounded taxonomy containing different identity cues~\cite{neplenbroek2025reading}. These types include, for example, cultural references, food, hobbies, and behavior descriptions. The demographic itself is never explicitly named in either the cued or the neutral condition. We examine three attribute conditions across race, gender, and age. Detailed categories include Black, Asian, and White for race, male and female for gender, and child, adolescent, adult, and older adult for age, resulting in nine conditions in total.

For each of the $n=50$ underlying scenarios we generate a matched set of $K=5$-turn user and assistant exchanges. We randomly sample cues from the pool with exactly one cue per user turn. Every context within a matched set shares the exact same topical scaffolding, which means that the topic of each assistant turn remains fixed across the conditions. From this shared backbone we branch out into the required test conditions. These include the \textit{identity-cued} variants for each demographic level, a \textit{neutral} baseline using identity-neutral fillers, and \textit{explicit} contexts that state the demographic directly for behavioral calibration. Each set concludes with the same query asking to recommend five movies and output only the title and year on each line. We obtain responses using greedy decoding ($T=0$). We focus on movies in the main paper and report results for books and articles in Appendix~\ref{sec:app-books-articles}.

For the intersectional experiments in \S\ref{sec:lin} and \S\ref{sec:abl}, we combine two identity dimensions in the same dialogue. We set $K=4$ with two cues from the first dimension and two from the second. This ensures the cross-dimension comparison stays fair. Each mixed-cue condition is accompanied by two dose-matched controls focusing on a single dimension. We construct these by replacing the alternate dimension's cues with neutral fillers. Finally, we randomize the order in which specific identity cues appear across samples to mitigate any positional confounding.


\subsection{Models}
\label{sec:models}

We evaluate five instruction-tuned LLMs ranging from 7B to 14B parameters. These include Llama-3-8B, Mistral-7B-v0.3, Qwen3-8B, Qwen3-14B, and Phi-4\footnote{We use \texttt{meta-llama/Meta-Llama-3-8B-Instruct} under the Meta Llama 3 Community License; \texttt{mistralai/Mistral-7B-Instruct-v0.3}, \texttt{Qwen/Qwen3-8B}, and \texttt{Qwen/Qwen3-14B} under the Apache License 2.0; and \texttt{microsoft/phi-4} under the MIT License.}. We run the correlation and intersectional experiments on all five models, and the ablation and capability-preservation experiments in \S\ref{sec:abl} on the three smaller models.

\subsection{Behavioral metrics}
\label{sec:eval-behavior}

We measure the personalization shift using two complementary metrics.

\paragraph{Semantic Embedding Distance on descriptions (SED-desc).}
For each sample, we resolve the five recommended titles to their TMDB\footnote{https://www.themoviedb.org/} plot overviews. We concatenate these overviews into a single string per condition and encode them using the pretrained sentence transformer \texttt{all-mpnet-base-v2}~\citep{reimers2019sentence}. SED-desc is one minus the cosine similarity of the two resulting embeddings. Higher values indicate a larger semantic shift between the cued and neutral recommendations. This continuous, taxonomy-free metric captures item substitution and semantic similarity between non-identical items. It also detects shifts in the overall thematic character of the recommendation set. 


\paragraph{Content Alignment Ratio (CAR).}
While SED-desc measures the magnitude of the shift, it does not measure its direction; a response can drift semantically without becoming more aligned with a stereotype. To capture this directional shift, we use GPT-4o to classify whether each recommended item is associated with the target identity using a pre-specified taxonomy (Appendix~\ref{gpt_4o}). We define the CAR as the fraction of the sample's five items tagged as identity-associated. For sample $i$ and target taxonomy $t$, the final CAR shift is calculated as $\mathrm{CAR}(C_{\mathrm{cue}}^{(i)};t)-\mathrm{CAR}(C_{\mathrm{neutral}}^{(i)};t)$. Scoring the paired neutral response against the same taxonomy controls for baseline stereotype content in the model's default output. To validate the automated labeling, a human annotator independently re-annotated a stratified sample of the recommended items using the same taxonomy. The human and GPT-4o achieved an overall agreement of 0.8, with disagreements primarily concentrated on the adult/neutral boundary, as shown in Appendix~\ref{gpt_4o}.

The two metrics answer different questions. SED-desc measures how far a response moved; CAR measures which identity-associated content it contains. 



\begin{table*}[t]
\centering \small
\setlength{\tabcolsep}{2pt}
\begin{tabular}{ll@{\hspace{6pt}}r@{}l@{\hspace{4pt}}r@{}l@{\hspace{6pt}}r@{}l@{\hspace{4pt}}r@{}l@{\hspace{6pt}}r@{}l@{\hspace{4pt}}r@{}l@{\hspace{6pt}}r@{}l@{\hspace{4pt}}r@{}l@{\hspace{6pt}}r@{}l@{\hspace{4pt}}r@{}l@{\hspace{6pt}}}
\toprule
\multicolumn{2}{c}{} & \multicolumn{4}{c}{Llama-3-8B} & \multicolumn{4}{c}{Mistral-7B} & \multicolumn{4}{c}{Qwen3-8B} & \multicolumn{4}{c}{Qwen3-14B} & \multicolumn{4}{c}{Phi-4} \\
\cmidrule(lr){3-6}\cmidrule(lr){7-10}\cmidrule(lr){11-14}\cmidrule(lr){15-18}\cmidrule(lr){19-22}
Dim.\ & Attribute & \multicolumn{2}{c}{SED} & \multicolumn{2}{c}{CAR} & \multicolumn{2}{c}{SED} & \multicolumn{2}{c}{CAR} & \multicolumn{2}{c}{SED} & \multicolumn{2}{c}{CAR} & \multicolumn{2}{c}{SED} & \multicolumn{2}{c}{CAR} & \multicolumn{2}{c}{SED} & \multicolumn{2}{c}{CAR} \\
\midrule
\multirow{3}{*}{race} & Black & +0.82 & \makebox[0.9em][l]{$^{***}$} & \textbf{+0.85} & \makebox[0.9em][l]{$^{***}$} & \textbf{+0.68} & \makebox[0.9em][l]{$^{***}$} & \textbf{+0.72} & \makebox[0.9em][l]{$^{***}$} & \textbf{+0.70} & \makebox[0.9em][l]{$^{***}$} & \textbf{+0.82} & \makebox[0.9em][l]{$^{***}$} & +0.49 & \makebox[0.9em][l]{$^{***}$} & \textbf{+0.81} & \makebox[0.9em][l]{$^{***}$} & \textbf{+0.64} & \makebox[0.9em][l]{$^{***}$} & \textbf{+0.75} & \makebox[0.9em][l]{$^{***}$} \\
 & Asian & \textbf{+0.87} & \makebox[0.9em][l]{$^{***}$} & +0.71 & \makebox[0.9em][l]{$^{***}$} & +0.51 & \makebox[0.9em][l]{$^{***}$} & +0.42 & \makebox[0.9em][l]{$^{**}$} & +0.38 & \makebox[0.9em][l]{$^{*}$} & +0.74 & \makebox[0.9em][l]{$^{***}$} & +0.26 & \makebox[0.9em][l]{} & +0.66 & \makebox[0.9em][l]{$^{***}$} & +0.26 & \makebox[0.9em][l]{} & +0.36 & \makebox[0.9em][l]{$^{*}$} \\
 & White & +0.34 & \makebox[0.9em][l]{$^{*}$} & -- & \makebox[0.9em][l]{} & +0.29 & \makebox[0.9em][l]{} & -- & \makebox[0.9em][l]{} & +0.18 & \makebox[0.9em][l]{} & -- & \makebox[0.9em][l]{} & -0.01 & \makebox[0.9em][l]{} & +0.18 & \makebox[0.9em][l]{} & +0.30 & \makebox[0.9em][l]{} & -- & \makebox[0.9em][l]{} \\
\midrule
\multirow{2}{*}{gender} & Male & +0.65 & \makebox[0.9em][l]{$^{***}$} & +0.58 & \makebox[0.9em][l]{$^{***}$} & +0.37 & \makebox[0.9em][l]{$^{*}$} & -0.16 & \makebox[0.9em][l]{} & +0.37 & \makebox[0.9em][l]{$^{*}$} & +0.61 & \makebox[0.9em][l]{$^{***}$} & +0.34 & \makebox[0.9em][l]{$^{*}$} & +0.32 & \makebox[0.9em][l]{$^{*}$} & +0.26 & \makebox[0.9em][l]{} & +0.40 & \makebox[0.9em][l]{$^{*}$} \\
 & Female & +0.56 & \makebox[0.9em][l]{$^{***}$} & +0.66 & \makebox[0.9em][l]{$^{***}$} & +0.11 & \makebox[0.9em][l]{} & -- & \makebox[0.9em][l]{} & +0.27 & \makebox[0.9em][l]{} & +0.66 & \makebox[0.9em][l]{$^{***}$} & +0.40 & \makebox[0.9em][l]{$^{**}$} & +0.69 & \makebox[0.9em][l]{$^{***}$} & +0.37 & \makebox[0.9em][l]{$^{*}$} & +0.26 & \makebox[0.9em][l]{} \\
\midrule
\multirow{4}{*}{age} & Child & +0.71 & \makebox[0.9em][l]{$^{***}$} & +0.65 & \makebox[0.9em][l]{$^{***}$} & +0.39 & \makebox[0.9em][l]{$^{**}$} & +0.49 & \makebox[0.9em][l]{$^{***}$} & +0.56 & \makebox[0.9em][l]{$^{***}$} & +0.80 & \makebox[0.9em][l]{$^{***}$} & \textbf{+0.52} & \makebox[0.9em][l]{$^{***}$} & +0.59 & \makebox[0.9em][l]{$^{***}$} & +0.62 & \makebox[0.9em][l]{$^{***}$} & +0.48 & \makebox[0.9em][l]{$^{**}$} \\
 & Adol.\ & +0.19 & \makebox[0.9em][l]{} & +0.58 & \makebox[0.9em][l]{$^{***}$} & +0.02 & \makebox[0.9em][l]{} & -- & \makebox[0.9em][l]{} & +0.29 & \makebox[0.9em][l]{} & -- & \makebox[0.9em][l]{} & +0.03 & \makebox[0.9em][l]{} & -- & \makebox[0.9em][l]{} & +0.07 & \makebox[0.9em][l]{} & -0.04 & \makebox[0.9em][l]{} \\
 & Adult & +0.17 & \makebox[0.9em][l]{} & +0.29 & \makebox[0.9em][l]{$^{*}$} & +0.08 & \makebox[0.9em][l]{} & -- & \makebox[0.9em][l]{} & +0.12 & \makebox[0.9em][l]{} & +0.11 & \makebox[0.9em][l]{} & +0.04 & \makebox[0.9em][l]{} & +0.06 & \makebox[0.9em][l]{} & -0.02 & \makebox[0.9em][l]{} & -0.02 & \makebox[0.9em][l]{} \\
 & Older & +0.32 & \makebox[0.9em][l]{$^{*}$} & +0.22 & \makebox[0.9em][l]{} & +0.63 & \makebox[0.9em][l]{$^{***}$} & +0.54 & \makebox[0.9em][l]{$^{***}$} & -0.04 & \makebox[0.9em][l]{} & -- & \makebox[0.9em][l]{} & +0.10 & \makebox[0.9em][l]{} & +0.48 & \makebox[0.9em][l]{$^{**}$} & +0.26 & \makebox[0.9em][l]{} & +0.27 & \makebox[0.9em][l]{} \\
\bottomrule
\end{tabular}

\caption{Per-sample Pearson $r$ between the normalized raw activation contrast magnitude $s_{\Delta A}$ at $L^\star$ and each
behavioral DV ($n=50$), for both DVs side by side: SED-desc
and CAR. \textbf{Bold} marks the strongest correlation in each (model, DV)
column. Stars are BH $q$ within model: $^*\,q<.05$, $^{**}\,q<.01$,
$^{***}\,q<.001$; -- marks a result whose paired
cue-minus-neutral CAR has zero variance or fewer than three nonzero samples.}

\label{tab:phase1}
\vspace{-10pt}
\end{table*}

\subsection{Internal representational quantity}

\label{sec:eval-deltaa}

Let $\mathbf{A}_L(C,x)$ denote the post-attention residual-stream activation at the final query-token position of layer $L$, given context $C$ and query $x$. We focus on the last-token activation because it already integrates every preceding turn. We define the raw contrast between the matched cued and neutral contexts of sample $i$ as
\[
\Delta\mathbf{A}_L^{(i)} = \mathbf{A}_L(C_{\mathrm{cue}}^{(i)},x) - \mathbf{A}_L(C_{\mathrm{neutral}}^{(i)},x).
\]
Adapting the contrast magnitude studied by \citet{dherin2025learning}, we compute the normalized magnitude of this contrast to test if it corresponds to the observed personalization effects:
\begin{equation}
s_{\Delta A}^{(i,L)} = \frac{\lVert \Delta\mathbf{A}_L^{(i)} \rVert_2}{\lVert \mathbf{A}_L(x) \rVert_2},
\label{eq:deltaa}
\end{equation}
where $\mathbf{A}_L(x)$ is the activation from a forward pass on the bare query alone. Because this denominator is identical across samples and conditions, it places layers of differing residual-stream scales on a common footing before averaging. We compute Equation~\eqref{eq:deltaa} using cached activations at the condition-specific layer set $L^\star$, identified via activation patching (Appendix~\ref{sec:app-patching}), and average across those layers. This approach directly connects to the foundational contrastive constructions used in activation steering \citep{zou2023representation, li2023inference, rimsky2024steering}, with targeted adaptations for our detection setting. While steering typically extracts a direction by contrasting a concept-bearing context against a bare query, we contrast against a matched neutral conversation to isolate the cue's specific effect from the general act of holding a dialogue. Furthermore, we use its magnitude as our detection signal.

\section{Experiments}
\label{sec:experiments}

This section evaluates the internal signal in three stages.
Section~\ref{sec:corr} tests whether its normalized magnitude tracks the personalization
behavior across models and conditions. Section~\ref{sec:lin} examines how
the model handles conversations that carry several overlapping identity
cues at once, comparing how the output and the internal
representation combine those cues. Section~\ref{sec:abl} then
projects the implicated direction out of the residual stream during
generation, testing whether a
single dimension's contribution can be removed on its own.

\subsection{Activation-Contrast Magnitude Tracks the Behavioral Shift}
\label{sec:corr}

We first replicate the established behavioral fact that implicit
demographic cues shift a model's recommendations
\citep{kotek2023gender, hofmann2024ai, neplenbroek2025reading} by quantifying it. 
We run the single-dimension setup on nine conditions across all five
models, with $n=50$ matched triplets per condition, and test each condition
for a significant personalization shift on either behavioral metric. The behavior is real but heterogeneous. A significant shift appears in 7 of 9 conditions on Llama, 6 on Mistral, 4 on Qwen3-8B, 7 on Qwen3-14B, and 7 on Phi-4. Two conditions act as universal positive anchors, shifting on all five models, the Black-cue and child-cue conditions, while the White-cue condition is a universal null with no significant shift on any model. The adult-cue condition is also null on Llama, Mistral, and Qwen3-8B but significant on Qwen3-14B and Phi-4. Per-condition effect sizes for all five models are mapped in Appendix~\ref{sec:app-phenom}.

For each model and condition, we then take $L^\star$, the layer set that
mediates the cue's effect on the output distribution, identified by the
activation patching detailed in Appendix~\ref{sec:app-patching}. We average
$s_{\Delta A}$ over $L^\star$ to obtain one value per sample, and correlate
it with each paired behavioral metric across the 50 samples. Through this
analysis, we find that the raw activation-contrast magnitude tracks the shift,
as shown in Table~\ref{tab:phase1}. This is most evident in the Black-cue and child-cue conditions, which produced a significant behavioral shift across all five models. In these strong conditions, the correlation is positive on every model: for the Black-cue condition, Pearson $r$ ranges from $0.49$ to $0.82$ against SED-desc and $0.72$ to $0.85$ against neutral-normalized CAR; for the child-cue condition, it ranges from $0.39$ to $0.71$ and $0.48$ to $0.80$. The largest single correlation is Llama's Asian-cue SED result at 0.87. To ensure statistical robustness, we applied Benjamini-Hochberg (BH) correction within each model.

Crucially, the strength of this internal tracking scales with the magnitude of the model's behavioral shift. Across all 45 model$\times$condition combinations, larger SED shifts correspond to significantly stronger correlations (Spearman $\rho=0.54$, $p < 0.001$). This same pattern holds for the 35 combinations with scorable paired CAR shifts (Spearman $\rho=0.57$, $p < 0.001$). Conversely, when the model's behavior does not shift, the internal correlation weakens sharply. For example, in the White-cue condition—which acts as a universal behavioral null across all models—SED correlations peak at only $0.34$, and CAR shifts are so negligible they are unscorable on four of the five models (with the remaining model showing a weak $r=0.18$), with details in Appendix~\ref{sec:app-corr-concord}.

\subsection{Multiple Cues: Linear Composition, an Interacting Behavior}
\label{sec:lin}

When two identity cues co-occur, we ask two questions at two levels: how the model's output combines the cues, and how the internal
direction that drives that output combines them. We answer both on
the dose-matched factorial design of Section~\ref{sec:dataset} over three dimension pairs: race$\times$gender, race$\times$age, gender$\times$age.

\paragraph{Behavior.}
We evaluate the behavioral shifts using three neutral-normalized CAR metrics: the individual tag rates for each dimension ($\mathrm{CAR}_A$, $\mathrm{CAR}_B$) and the rate of items carrying stereotype labels from both dimensions at once ($\mathrm{CAR}_{\mathrm{both}}$). We use these to examine whether one identity cue modulates the expression of another, and whether their joint effect is simply additive.

\begin{figure}[t]
\centering
\includegraphics[width=\linewidth]{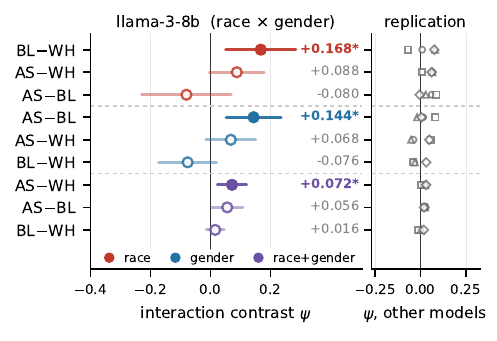}
\caption{Interaction contrasts $\psi$ on race$\times$gender for Llama-3-8B
(left) and the other four models (right). Each row is one $2{\times}2$
contrast: the row label gives the race difference (BL, AS and WH are Black,
Asian and White), each taken against the same female-minus-male gender
difference. 
Right-panel markers are Mistral-7B ($\circ$), Qwen3-8B ($\square$),
Qwen3-14B ($\triangle$) and Phi-4 ($\Diamond$).
}
\label{fig:psi}
\vspace{-10pt}
\end{figure}

To measure cross-cue modulation, we use the difference-of-differences contrast $\psi$ \cite{maxwell2017designing}, which isolates how much a second identity cue amplifies or suppresses the behavioral shift caused by the first. Independent cues would yield $\psi=0$. Figure~\ref{fig:psi} maps these contrasts for race$\times$gender. On Llama-3-8B, we observe a significant interaction where race content dominates at the expense of gender: cueing Black rather than White raises the share of race-stereotyped movies by $+0.26$ under a female cue, but only by $+0.10$ under a male cue ($\psi=+0.168$, $q=0.021$). Concurrently, gender expression drops in this intersection, with the Asian-female condition yielding significantly more gender-stereotyped movies than the Black-female condition while the two barely differ under a male cue ($\psi=+0.144$, $q=0.010$). A third contrast captures the co-expression of both tags ($\psi=+0.072$, $q=0.016$). Overall, this reshaping is sparse and model-specific; across all canonical runs, only 12 contrasts survive BH correction, with full results in Appendix~\ref{sec:app-int-def}. 

Furthermore, the joint behavioral shift is strictly sub-additive. When both cues are present, the resulting personalization is reliably smaller than the sum of the individual single-cue shifts. Specifically, when evaluating the paired semantic shift (SED-desc), the observed mixed shift falls $28\%$ to $38\%$ below the strict additive prediction across the tested models (Appendix~\ref{sec:app-interaction}). This compression also holds for the joint-tag rate $\mathrm{CAR}_{\mathrm{both}}$, and a strict test finds no super-additive amplification almost anywhere. Therefore, the behavioral output combines its cues sub-additively, reshaping them in a sparse, model-specific way. 

\begin{figure}[t]
\centering
\includegraphics[width=\linewidth]{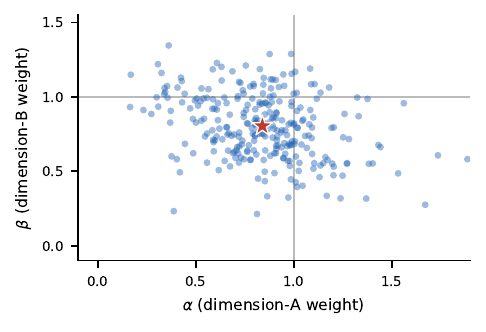}
\caption{Per-sample decomposition weights on Llama race$\times$gender,
pooled over six mixed-cue conditions ($n=278$; 22 structurally invalid generated
dialogues are excluded). Each dot is one sample's fitted
$(\alpha,\beta)$ on raw $\Delta\mathbf A$ at $L^\star$; the red star is
the grand mean $(.84,.80)$ and the thin lines mark exact addition $(1,1)$.
}
\label{fig:alphabeta}
\vspace{-10pt}
\end{figure}

\paragraph{Mechanism.}
To understand how internal representations combine, we analyze the raw contrast vector $\Delta\mathbf{A}$. For each sample and selected layer, let $\mathbf{v}_A$, $\mathbf{v}_B$, and $\mathbf{v}_{AB}$ denote the cue-minus-neutral contrast vectors for the two dose-matched single-cue contexts and their mixed-cue equivalent. If the model simply added the single-cue directions, we would observe $\mathbf{v}_{AB} = \mathbf{v}_A + \mathbf{v}_B$. We test this additive prediction by fitting $\mathbf{v}_{AB} \approx \alpha\mathbf{v}_A + \beta\mathbf{v}_B$ per sample and layer. The residual fit $\rho = 1 - \lVert\mathbf{v}_{AB} - \hat{\mathbf{v}}_{AB}\rVert_2 / \lVert\mathbf{v}_{AB}\rVert_2$ measures how much of the mixed vector is reconstructed by the two-vector span, while the weights $(\alpha, \beta)$ describe the mixture.  

Figure~\ref{fig:alphabeta} illustrates this decomposition for the Llama race$\times$gender condition. The grand mean of the weights is $(0.84, 0.80)$, which sits below the exact $(1,1)$ additive prediction. Across all 15 model and pair combinations, the mean residual fit $\rho$ ranges from $0.576$ to $0.704$. Reconstructing the mixed vector using both single-cue components achieves an average fit of $0.646$. This outperforms attempting to explain the mixed state using only a single demographic direction, which averages a fit of only $0.443$. In fact, the two-cue combination improves upon the best single-cue fit across all evaluated samples. This indicates that when two cues are present, the model actively integrates both identity directions into its internal representation rather than dropping one. Crucially, this linear structure is highly sample-specific: substituting a basis from a different sample in the same condition collapses the mean fit to $0.095$. 

Because the fit is sample-specific, this averaged vector retains only the component shared across samples. The next section tests whether ablating that shared component alone suffices to control the behavior.

\begin{table*}[t]
\centering \footnotesize
\setlength{\tabcolsep}{4pt}
\begin{tabular}{ll rrrr rrrr}
\toprule
& & \multicolumn{4}{c}{mistral-7b} & \multicolumn{4}{c}{qwen3-8b} \\
\cmidrule(lr){3-6}\cmidrule(lr){7-10}
Pair & Target & $\Delta_{\rm pure(t)}$ & Cohen's $d$ & Prompt~$\Delta$ & Ratio & $\Delta_{\rm pure(t)}$ & Cohen's $d$ & Prompt~$\Delta$ & Ratio \\
\midrule
\multirow{2}{*}{race\,$\times$\,gender} & race   & $+0.226^{***}$ & $+1.59$ & $-0.008$ & $\dagger$ & $+0.176^{***}$ & $+0.72$ & $+0.027$ & \textbf{6.6$\times$} \\
 & gender & $+0.235^{***}$ & $+1.85$ & $+0.014$ & \textbf{16.7$\times$} & $+0.078^{***}$ & $+0.41$ & $+0.031$ & \textbf{2.5$\times$} \\
\addlinespace[1pt]
\multirow{2}{*}{race\,$\times$\,age}    & race   & $+0.154^{***}$ & $+1.22$ & $-0.005$ & $\dagger$ & $+0.306^{***}$ & $+1.04$ & $+0.052$ & \textbf{5.9$\times$} \\
 & age    & $+0.169^{***}$ & $+1.13$ & $+0.007$ & \textbf{23.3$\times$} & $+0.098^{***}$ & $+0.41$ & $+0.020$ & \textbf{4.8$\times$} \\
\addlinespace[1pt]
\multirow{2}{*}{gender\,$\times$\,age}  & gender & $+0.029^{***}$ & $+0.26$ & $-0.013$ & $\dagger$ & $+0.078^{***}$ & $+0.42$ & $+0.028$ & \textbf{2.8$\times$} \\
 & age    & $-0.008$       & $-0.08$ & $+0.004$ & -- & $+0.055^{***}$ & $+0.24$ & $+0.006$ & \textbf{9.2$\times$} \\
\bottomrule
\end{tabular}

\caption{Direction ablation at $\alpha=5$ for Mistral-7B and Qwen3-8B,
measured on SED-desc.
$\Delta_{\mathrm{pure}(t)}$ is the condition-specific shift away from the reference response. Ratio is
ablation over prompt; $\dagger$ marks entries where the prompt moved in the
wrong direction, and -- marks the null-ablation result. Stars: $^*\,p<.05$,
$^{**}\,p<.01$, $^{***}\,p<.001$. 
}
\label{tab:ablation}
\vspace{-6pt}
\end{table*}

\begin{figure}[t]
\centering
\includegraphics[width=\linewidth]{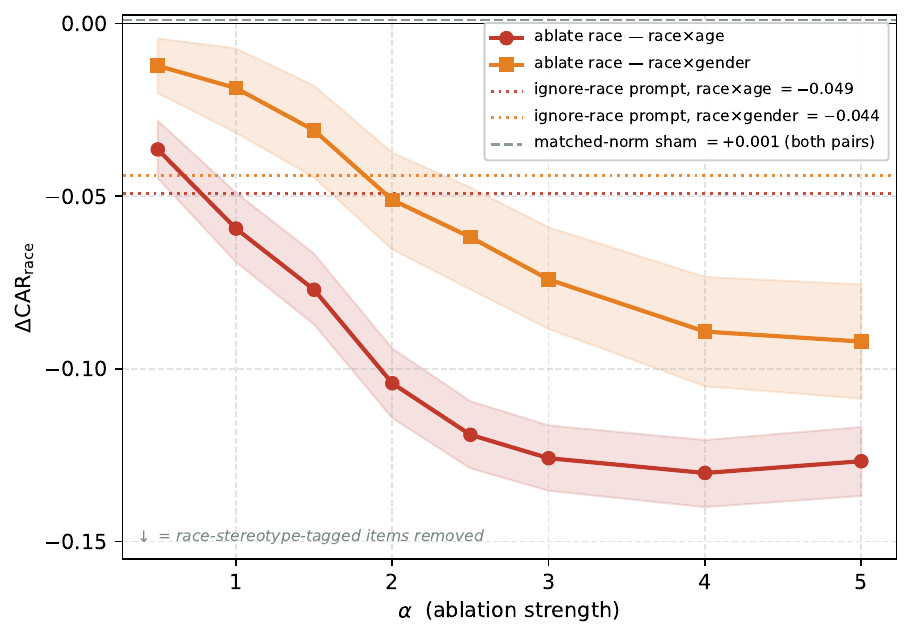}
\caption{Removes the race-associated direction in two Qwen3-8B multi-cue contexts. The reduction increases with the intervention strength $\alpha$ in both race$\times$age and race$\times$gender contexts, with a negative $\Delta\mathrm{CAR}_{\mathrm{race}}$ indicating fewer race-tagged recommendations. The ``ignore race'' prompt baselines are shown by dotted lines, while the dashed line represents the original random-direction sham. The bands indicate bootstrap $95\%$ CIs.}
\label{fig:carresp}
\vspace{-10pt}
\end{figure}



\begin{table}[t]
\centering
\small
\setlength{\tabcolsep}{4pt}
\begin{tabular}{ll l rr}
\toprule
Model & Pair & target & $\Delta_{\rm CAR_T}$ & $\Delta_{\rm CAR_O}$ \\
\midrule
\multirow{6}{*}{Llama}
 & \multirow{2}{*}{\shortstack[l]{race\\$\times$\,gender}} & race & $-.030^{*}$ & $+.002$ \\
 &  & gender & $+.014$ & $-.002$ \\
\addlinespace[1pt]
 & \multirow{2}{*}{\shortstack[l]{race\\$\times$\,age}} & race & $-.009$ & $+.001$ \\
 &  & age & $+.018$ & $-.026^{***}$ \\
\addlinespace[1pt]
 & \multirow{2}{*}{\shortstack[l]{gender\\$\times$\,age}} & gender & $+.006$ & $-.002$ \\
 &  & age & $+.002$ & $-.028^{***}$ \\
\midrule
\multirow{6}{*}{Mistral}
 & \multirow{2}{*}{\shortstack[l]{race\\$\times$\,gender}} & race & $+.006$ & $-.009$ \\
 &  & gender & $-.169^{***}$ & $+.004$ \\
\addlinespace[1pt]
 & \multirow{2}{*}{\shortstack[l]{race\\$\times$\,age}} & race & $+.001$ & $-.002$ \\
 &  & age & $-.002$ & $+.006$ \\
\addlinespace[1pt]
 & \multirow{2}{*}{\shortstack[l]{gender\\$\times$\,age}} & gender & $-.039^{***}$ & $-.003$ \\
 &  & age & $-.024^{***}$ & $+.027^{***}$ \\
\bottomrule
\end{tabular}

\caption{CAR change for the removed dimension ($\Delta_{\rm CAR_T}$) and for
the co-present one ($\Delta_{\rm CAR_O}$); negative means fewer tagged
recommendations. Each row uses the strength
maximizing the target reduction, listed with the full grid in
Appendix~\ref{sec:app-ablation-grids}. Stars: one-sided Wilcoxon for the
target, two-sided for the co-present dimension.}
\label{tab:selective-car}
\vspace{-10pt}
\end{table}

\subsection{Causal Control via Direction Ablation}
\label{sec:abl}
We test whether one cue's contribution can be selectively removed and how this affects a co-present cue. For each dimension, we average the raw single-cue activation contrasts across samples~\cite{arditi2024refusal, zou2023representation}, remove the component along the partner dimension, and unit-normalize the result as $\hat{v}_{\mathrm{dim}}$. At the union of the two dimensions' top two $L^\star$ layers, we apply directional ablation using a forward pre-hook on the post-attention layernorm:
\begin{equation}
h' = \left(I - \alpha\, \hat v_{\dim}\hat v_{\dim}^{\top}\right) h
\label{eq:ablation}
\end{equation}

where $h$ is the incoming activation and $\alpha$ dictates the intervention strength. We sweep $\alpha \in [0.5, 5]$ (unrelated to the decomposition weights), where $\alpha=1$ deletes the component exactly and $\alpha>1$ continues past it, across Llama-3-8B, Mistral-7B, and Qwen3-8B, comparing this internal ablation against random-direction shams and an explicit ``ignore demographic'' prompt.  

As Figure~\ref{fig:carresp} shows, removing the race-associated direction steadily reduces race-tagged recommendations as $\alpha$ increases, dropping $\Delta\mathrm{CAR}_{\mathrm{race}}$ by $0.128$ in the race$\times$age context and $0.092$ in the race$\times$gender context at $\alpha=5$. In contrast, a matched-norm sham produces minimal change on the taxonomy metric (+0.001), confirming the suppression is driven by the specific learned direction rather than generic perturbation. Across the full results on SED, as shown in Table~\ref{tab:ablation}, internal ablation outperforms the explicit ``ignore demographic'' prompt by up to $9.2\times$ on Qwen3-8B, while the prompt occasionally backfires and moves outputs in the wrong direction on Mistral. Throughout this sweep, general capabilities remain robust: MMLU~\cite{hendrycks2020measuring} accuracy drops by at most $2.8$ points even at the maximum strength of $\alpha=5$, with full results in Appendix~\ref{sec:app-mmlu}.

A stricter test is whether this targeted removal successfully spares the co-present dimension, which we evaluate by tracking both the targeted and co-present taxonomy counts simultaneously, as shown in Table~\ref{tab:selective-car}. Of the 18 settings, targeted taxonomy-tagged content falls significantly in seven, and in three of those, the co-present dimension remains entirely unchanged. The cleanest example is Mistral's race$\times$gender context: removing the gender direction lowers gender-tagged recommendations by $0.169$, while race-tagged content moves by only $+0.004$ across all intervention strengths. Llama race$\times$gender and Mistral gender$\times$age exhibit similar selectivity. However, this decoupling is not automatic; on Qwen3-8B race$\times$age, both rates fall together ($-0.130$ race, $-0.104$ age). Because generic semantic drift affects both references, these selective edits can only be adjudicated using our taxonomy counts (Appendix~\ref{sec:app-orth}).

\section{Discussion}
\paragraph{Linking an internal representation to a behavior.}
While the behavioral side of implicit personalization is well documented \citep{neplenbroek2025reading, hofmann2024ai, jin2024implicit}, the link to the model's internal computation has remained missing. We bridge this gap by demonstrating that the magnitude of the cue-induced contrast at specific layers directly tracks how strongly a model personalizes. This correlation reaches up to $r=0.87$ in the strongest conditions, and falls a lot in conditions where no shift occurs. Because our experimental design compares matched neutral dialogues and uses dose-matched controls, we attribute this tracking to the cue rather than the surrounding conversation, though not to demographic content specifically rather than the topical content correlated with it.
Importantly, the chain we test runs strictly from cue to activation to behavior. This isolates the mechanism of implicit personalization without claiming that the model explicitly infers a user's demographic identity beforehand.


\paragraph{Linear geometry versus non-linear behavior.}
We observe a distinct dissociation between internal geometry and external behavior. Internally, the mixed-cue contrast actively integrates both identity directions, forming a linear structure well reconstructed from the sample's \emph{own} single-cue components ($\bar\rho=0.576$--$0.704$); substituting a different sample's basis recovers only $0.038$ to $0.156$. Externally, however, the behavior compresses rather than adding, falling $28\%$ to $38\%$ below the sum of the single-cue shifts. While concurrent work reports similar dissociations \citep{tripathy2026fair}, our dose-matched design quantifies both sides. Because the internal and external metrics are not directly commensurate, this experiment leaves open whether the nonlinearity originates in the unexplained internal residual, the downstream readout, or both. 

\paragraph{Per-dimension control is model-specific.}
Projecting out one sample-averaged direction per cue dimension can selectively suppress one component of a multi-cue response, but only for some models and dimensions. On Mistral, ablating gender lowers gender-tagged content by up to $0.169$ while leaving race-tagged content unchanged; on Llama, ablating race leaves gender-tagged content unchanged. Qwen3-8B shows less separation: in its clearest case, race- and age-tagged content fall together. On SED-desc, the intervention is more consistent: at $\alpha=5$, it suppresses all six Qwen3-8B targets and exceeds the ignore-demographics prompt by up to $9.2\times$, whereas on Mistral the prompt moves one target in each pair toward the cue. At exact projection, MMLU remains within $0.6$ points of baseline. These results extend direction-based control beyond stated-attribute settings \citep{karvonen2025robustly, kantharuban2025stereotype} to implicit, multi-cue contexts, but show that suppression is more reliable than selectivity.


\section{Conclusion}
In this study, we connect implicit personalization to a locatable internal signal: the cue-induced activation contrast at targeted layers. First, its normalized magnitude tracks a model's behavioral shift, reaching $r=0.87$ in the strongest conditions. Second, when multiple cues co-occur, internal representations exhibit a substantial linear structure, whereas outward behavior is strictly sub-additive. Third, projecting out a sample-averaged direction successfully suppresses targeted stereotype content, outperforming explicit prompting while preserving general capabilities, though selectivity remains model-specific. Grounding implicit personalization in a causally validated activation contrast transforms it from a phenomenon observed only at the output into an internal mechanism that can be inspected, decomposed, and causally controlled.

\section*{Limitations}

Our evaluation scope is primarily bounded to the movie recommendation
domain. Books and articles are tested only in the single-dimension setting
and only on Llama (Appendix~\ref{sec:app-books-articles}). Furthermore, the
direction ablation covers three $7$ to $8$B instruction-tuned models, with
the two $14$B models contributing correlation and composition evidence
only (\S~\ref{sec:models}). Generalization to larger models and
base non-instruct models remains open. 

Construct validity presents a second limitation. While our cued and
neutral contexts share identical scaffolding, the cue phrases themselves
carry distinct semantic content. A demographic cue intrinsically carries
topical content and lexical specificity. As \citet{neplenbroek2026topics}
show, conversation topic can dominate output shifts in real
conversations. Accordingly, our design does not separate direct semantic
continuation from a more abstract demographic representation. Additionally, CAR
depends entirely on a fixed stereotype taxonomy. We mitigate this by
pairing it throughout with the taxonomy-free SED-desc and relying
on their agreement rather than either metric alone.

Mechanistically, the intersectional interaction we observe is
model-specific and does not survive correction when pooled across models,
making it a directional trend rather than a universal mechanism. The
selectivity of the internal ablation is model-specific. Ablating one
dimension's direction leaves the other dimension's tagged content intact on
Mistral and Llama but not on Qwen3-8B, and semantic distance moves away from
both pure targets in every case (Appendix~\ref{sec:app-orth}). This matches concept-erasure predictions for
rank-one removal of distributed attributes \citep{ravfogel2020null};
closed-form erasure like LEACE \citep{belrose2023leace} is a stronger
natural baseline left for future work. Finally, our metrics measure
expressed content. They cannot distinguish true geometric removal from
behavioral concealment \citep{gonen2019lipstick}, and we do not test
whether a linear probe could still recover the demographic from the
post-ablation residual stream.

\section*{Ethical considerations}

We adopt the operational definition of bias used in prior
implicit-personalization audits, namely a systematic shift in output
content conditioned on identity-correlated cues, and we follow
\citet{blodgett2020language} in stating that this is a
representational-harm framing rather than a general theory of bias. All
dialogues are synthetic GPT-4o output and contain no real individuals'
data, so no anonymization was required. The corpus is stereotype-laden by
construction, since eliciting stereotype-aligned content is the object of
study, and we did not filter it for offensive content for that reason. Our
use of GPT-4o infers demographic associations for synthetic dialogues and
publicly released media titles, never for real people. 

Our taxonomy operationalizes race as three categories, gender as two, and age as
four. These are coarse, non-exhaustive proxies chosen to match prior
implicit-personalization audits~\cite{neplenbroek2025reading}, not a claim about how identity is constituted.
Scoring stereotype alignment against a fixed lexicon also encodes a particular
cultural vantage point, so results should be read as evidence about model
behavior under this specific operationalization rather than about identity
groups themselves. All dialogues and recommendations are in English, so we cannot say whether the
same activation signature governs implicit personalization in other languages.

This work studies an existing bias and offers a means to detect and suppress
it. However, not every age-conditioned adaptation is harmful. For a child-cued user, age-appropriate recommendations may provide beneficial or protective accommodation; suppressing the age-conditioned direction could therefore remove useful adaptation together with stereotyping. We treat the ablation as a causal diagnostic rather than a deployment-ready fairness intervention. The same direction we ablate could in principle be \emph{added} to
amplify a demographic association---a dual-use risk shared with all
activation-steering research \citep{zou2023representation, li2025fairsteer}. We will release the evaluation and analysis code upon publication. We will not release the raw cue-construction templates because they operationalize demographic inference from indirect proxies; consequently, the synthetic dialogues cannot be regenerated exactly from the released artifacts.



\bibliography{custom}

\clearpage

\appendix
\section{Books and Articles Recommendation Queries}
\label{sec:app-books-articles}

The main paper fixes the query to movie recommendation. We additionally
run the single-dimension correlation pipeline on books and articles,
changing only the final query while reusing the same $n=50$ matched
dialogues and cue taxonomy. These supplementary domain checks use
Llama-3-8B only; all five models in the main experiment and every
multi-cue and intervention result use movies.

\paragraph{Setup.}
For each query and condition, we re-derive $L^\star$ with the same
activation-patching procedure (Appendix~\ref{sec:app-patching}) and
recapture the raw predictor
$\operatorname{mean}_{L\in L^\star}\lVert A_L(C_{\rm cue})-
A_L(C_{\rm neutral})\rVert_2/\lVert A_L(x)\rVert_2$ at the final prompt
token. CAR is paired by sample and subtracts the neutral response scored
against the same condition's taxonomy. White-cue CAR is unscorable in all
three domains after this subtraction.

The legacy books/articles runs use a paired response/title embedding
distance, not the TMDB-description SED used by the current movie run. The
table labels this as SED-response and we do not compare its absolute
magnitude with movie SED-description. CAR is defined identically and is
directly comparable across columns.

\paragraph{Result.}
The tracking pattern generalizes, but not uniformly. Averaged over the
eight shared conditions, raw-$\Delta A$ correlations are $.560$, $.458$,
and $.536$ on the available SED metric for movies, articles, and books;
mean corrected-CAR correlations are $.606$, $.387$, and $.547$ over seven
scorable cells per domain. Strong anchors persist: Black-cue CAR tracks at
$r=.85/.78/.67$ and child-cue CAR at $r=.65/.36/.82$ across
movies/articles/books. Domain heterogeneity remains visible: female-cue
CAR is strong on movies ($.66$) but null on articles ($-.10$) and books
($.17$). Thus the extra queries support transfer of the tracking pattern,
not a claim that every cue behaves identically in every domain.

\begin{table*}[t]
\centering
\footnotesize
\begin{tabular}{l cc cc cc}
\toprule
& \multicolumn{2}{c}{Movies} & \multicolumn{2}{c}{Articles} & \multicolumn{2}{c}{Books} \\
\cmidrule(lr){2-3}\cmidrule(lr){4-5}\cmidrule(lr){6-7}
Condition & SED-desc & CAR & SED-response & CAR & SED-response & CAR \\
\midrule
race: Black & $+0.82^{***}$ & $+0.85^{***}$ & $+0.68^{***}$ & $+0.78^{***}$ & $+0.76^{***}$ & $+0.67^{***}$ \\
race: Asian & $+0.87^{***}$ & $+0.71^{***}$ & $+0.62^{***}$ & $+0.61^{***}$ & $+0.61^{***}$ & $+0.87^{***}$ \\
race: White & $+0.34^{*}$ & -- & $+0.55^{***}$ & -- & $+0.33^{*}$ & -- \\
gender: male & $+0.65^{***}$ & $+0.58^{***}$ & $+0.56^{***}$ & $+0.62^{***}$ & $+0.52^{***}$ & $+0.49^{***}$ \\
gender: female & $+0.56^{***}$ & $+0.66^{***}$ & $+0.18$ & $-0.10$ & $+0.36^{*}$ & $+0.17$ \\
age: child & $+0.71^{***}$ & $+0.65^{***}$ & $+0.48^{***}$ & $+0.36^{**}$ & $+0.70^{***}$ & $+0.82^{***}$ \\
age: adolescent & $+0.19$ & $+0.58^{***}$ & $+0.32^{*}$ & $+0.11$ & $+0.49^{***}$ & $+0.46^{***}$ \\
age: older adult & $+0.32^{*}$ & $+0.22$ & $+0.28^{*}$ & $+0.34^{*}$ & $+0.51^{***}$ & $+0.34^{*}$ \\
\bottomrule
\end{tabular}

\caption{Per-sample Pearson $r$ between raw activation-contrast magnitude
at query-specific $L^\star$ and each behavioral DV on Llama-3-8B
($n=50$). CAR is cue minus the paired neutral response scored against the
same taxonomy. Movies use SED-description; the legacy articles/books runs
use paired response/title SED. Stars are uncorrected two-sided values:
$^{*}p<.05$, $^{**}p<.01$, $^{***}p<.001$; -- marks fewer than three
nonzero CAR shifts or zero variance. Adult was not scored in the two
legacy domains.}
\label{tab:books-articles}
\end{table*}

\section{GPT-4o Annotation Procedure for Content Alignment Ratio (CAR)}
\label{gpt_4o}

To compute the Content Alignment Ratio (CAR), each unique item
recommended by the model (e.g., a movie, book, or article title) was
annotated by GPT-4o (temperature $= 0$,
\texttt{response\_format: json\_object}) for its primary cultural
association along the relevant demographic dimension. Annotation was
performed item-by-item (chunk size $= 1$) to maximize label accuracy.
The annotator was given the following prompt:

\begin{quote}
\textit{You are an expert annotator for AI bias research.}

\textit{Task: For each recommended item, assign the identity label that
best describes its primary cultural association along the [dimension]
dimension.}

\textit{Valid labels: [e.g., ``female'', ``male'', ``neutral'']}\\
\textit{Definitions:}\\
\textit{-- ``[condition]'': [taxonomy description]}\\
\textit{-- ``neutral'': does not strongly fit any of the
identity-specific definitions above}

\textit{Items to annotate: [item]}\\
\textit{Instructions:}\\
\textit{1. Assign exactly one label from the valid label set.}\\
\textit{2. Choose ``neutral'' if the item does not clearly fit any
identity-specific definition.}\\
\textit{3. Apply the definitions strictly and independently.}

\textit{Return ONLY a valid JSON object with the format:}
\texttt{\{"items": [\{"item": "...", "label": "...", "reason":
"..."\}]\}}
\end{quote}

Taxonomy definitions were condition- and query-specific. For example,
for the \textbf{race $\times$ movies} condition, \textit{black} was
defined as ``movies associated with Black culture or featuring Black
experiences, including but not limited to: films by Spike Lee, Jordan
Peele, Tyler Perry, Ryan Coogler; films like Black Panther, Coming to
America, Boyz n the Hood, Get Out, Moonlight, The Color Purple,
Friday''; \textit{asian} as ``movies associated with Asian culture,
including but not limited to: anime films (Studio Ghibli, Your Name),
Korean cinema (Parasite, Oldboy), Bollywood films, Hong Kong action
films, films like Crazy Rich Asians, Everything Everywhere All at Once,
The Joy Luck Club''; and \textit{white} as ``movies strongly associated
with white American cultural stereotypes, including but not limited to:
Western films, country music biopics, films about rural or small-town
white America, films like Sweet Home Alabama, Friday Night Lights;
European historical dramas.''

Analogous definitions were applied for gender
(\textit{female}/\textit{male}) and age
(\textit{child}/\textit{adolescent}/\textit{adult}/\textit{older
adult}) across all query types (books, articles, movies). The CAR for a given condition was
then computed as the proportion of that condition's recommended items
annotated with the matching identity label.

For every single-dimension analysis, the behavioral variable is the
\emph{paired shift}, not the cued rate alone. For sample $i$ and condition
$g$, we compute
\[
\Delta\mathrm{CAR}_{i,g}=
\mathrm{CAR}(R^{\mathrm{cue}}_i;\mathcal T_g)-
\mathrm{CAR}(R^{\mathrm{neutral}}_i;\mathcal T_g),
\]
where the same condition taxonomy $\mathcal T_g$ scores both responses.
This subtraction removes stereotype-associated content already present in
the model's neutral output. A Table~\ref{tab:phase1} CAR cell is marked --
when the resulting vector has zero variance or fewer than three nonzero
sample shifts.

\paragraph{Human-annotation agreement.}
Because the dialogues and the CAR labels both come from GPT-4o, the
labels could in principle reflect that model's own priors rather than the
stated taxonomy. We therefore re-annotated a stratified sample of the
items actually recommended in our experiments---30 per dimension, drawn
across all identity levels and neutral---with a human annotator who was
given the same taxonomy definitions and was blind to the original labels.
Agreement with the model-based labels is 93\% on race, 83\% on gender,
and 63\% on age, or 80\% pooled over the 90 items. The disagreements are
structured rather than random: on race they occur only at the
\textit{white} boundary (European period
dramas), on gender only at the \textit{female}/neutral boundary, and on
age all eleven fall on the \textit{adult}/neutral boundary, where one
annotator reads prestige and relationship dramas as adult-associated and
the other as unmarked. The disputed adult condition is also among the
weakest behavioral conditions (\S\ref{sec:corr}), while the race labels that
carry the strongest results are the most reproducible. This comparison
tests whether a human can apply the stated taxonomy consistently with the
model-based labels. These percentages are agreement rates, not accuracy,
because neither label set is ground truth. The per-item labels are released
with the code.

\paragraph{Annotator details.} The human re-annotation was performed by one
of the authors, a US-based graduate student researcher,
who received the identical taxonomy definitions given to GPT-4o
(reproduced above) and was blind to the model-assigned labels. No
compensation was involved and no participants were recruited, so no
consent procedure applied. The task involves annotating publicly released
media titles rather than human subject data, and was therefore determined
exempt from ethics board review. Because the labels come from a single
annotator, we report raw agreement with the model-based labels rather than
inter-annotator reliability.        
\section{Activation Patching for $L^\star$ Selection}
\label{sec:app-patching}

The layer set $L^\star$ used in \S\ref{sec:eval-deltaa} (definition of
$\Delta\mathbf A$), \S\ref{sec:corr} (correlation), and \S\ref{sec:abl}
(direction ablation) is identified by post-attention activation
patching in the causal mediation tradition of
\citet{vig2020investigating, meng2022locating, zhang2024towards}. This
appendix specifies the procedure, the divergence measure, the
normalized effect statistic, and the hyperparameter choices.

\subsection{Post-attention patching at the last-token position}
\label{sec:app-patching-procedure}

For each model, each matched triplet $(C_{\mathrm{id}}, C_{\mathrm{neu}},
x)$, and each transformer layer $\ell \in \{0, \ldots, N_L - 1\}$
(where $N_L$ is the model's total layer count, see
\S\ref{sec:models}), we run three forward passes:
\begin{enumerate}\itemsep2pt
\item An \textbf{identity forward pass} on $C_{\mathrm{id}} + x$,
caching the post-attention residual-stream vector
$\mathbf{h}_{\mathrm{id}}^{(\ell)}$ at the last-token position.
\item A \textbf{neutral forward pass} on $C_{\mathrm{neu}} + x$,
caching $\mathbf{h}_{\mathrm{neu}}^{(\ell)}$ at the same position.
\item A \textbf{patched forward pass}: process $C_{\mathrm{neu}} + x$
but replace the post-attention hidden state at layer $\ell$, last-token
position, with $\mathbf{h}_{\mathrm{id}}^{(\ell)}$. Every other layer
runs normally.
\end{enumerate}
We patch only the last-token position $\mathbf{h}[:, -1, :]$ rather
than the full sequence. That position is where the first output token
is generated, and self-attention across earlier layers has already
mixed information from the entire prompt (including identity cues)
into it. Patching only this position avoids position-mismatch issues
between identity and neutral prompts that may have different token
counts, and aligns exactly with the position used to compute $\Delta\mathbf A$
in \S\ref{sec:eval-deltaa}.

\subsection{Jensen--Shannon divergence of next-token distributions}
\label{sec:app-patching-jsd}

Rather than generating full responses at every layer---which would
require expensive autoregressive decoding at each of the $N_L$ layers
per sample---we measure the effect of patching on the next-token
probability distribution. Let $p_{\mathrm{id}}(t) = P(y_1 = t \mid
C_{\mathrm{id}}, x)$ and $p_{\mathrm{neu}}(t) = P(y_1 = t \mid
C_{\mathrm{neu}}, x)$ be the next-token distributions over the
model's vocabulary $\mathcal{V}$ under the identity and neutral
prompts. Following \citet{lin1991divergence}, the per-sample baseline
divergence is defined as:

$$\begin{aligned} D_i &= \mathrm{JSD}(p_{\mathrm{id}}, p_{\mathrm{neu}}) \\     &= \tfrac{1}{2} D_{\mathrm{KL}}(p_{\mathrm{id}} \Vert{} m) + \tfrac{1}{2} D_{\mathrm{KL}}(p_{\mathrm{neu}} \Vert{} m), \end{aligned}$$

where $m = \tfrac{1}{2}(p_{\mathrm{id}} + p_{\mathrm{neu}})$ is the mixture distribution and the divergence is defined as
$$D_{\mathrm{KL}}(p \Vert{} q) = \sum_{t \in \mathcal{V}} p(t) \log \frac{p(t)}{q(t)}.$$

We use JSD \citep{lin1991divergence} rather than KL because JSD is
symmetric, bounded in $[0, \ln 2]$, and numerically stable when the
two distributions have limited overlap. This last property is important in our setting:
identity-specific tokens (e.g., the first subword of a
culturally-specific movie title) may carry substantial probability
under one condition but near-zero probability under the other, a
regime in which KL can produce arbitrarily large values while JSD
remains well-behaved. Empirically, the divergence concentrates on a
small number of semantically meaningful tokens: on inspected
high-divergence samples, the top 20 tokens (out of vocabularies of
33K--152K depending on model) account for over 90\% of total JSD,
and these tokens correspond to the first subwords of
identity-associated content items rather than generic function words.

\subsection{Normalized indirect effect}
\label{sec:app-patching-nie}

For each layer $\ell$ and sample $i$, we quantify the fraction of the
identity-vs-neutral output divergence that is causally mediated by
layer $\ell$'s post-attention representation. Following the causal
mediation framework of \citet{vig2020investigating}, we define the normalized
indirect effect (NIE) as:

$$\begin{aligned} \mathrm{NIE}(i, \ell) &= \operatorname{clip}\!\left( \right. \\ &\quad \left. 1 - \frac{\mathrm{JSD}\!\left(p_{\mathrm{patched}}^{(\ell)},\, p_{\mathrm{id}}\right)}{D_i + \epsilon},\; 0,\; 1 \right), \end{aligned}$$

where $p_{\mathrm{patched}}^{(\ell)}$ is the output distribution from
the patched forward pass at layer $\ell$, $D_i$ is the baseline
JSD from the previous section, and $\epsilon = 0.005$ prevents
numerical instability for samples with negligible baseline divergence.

The clipping range $[0, 1]$ interprets the score as a fraction of the
baseline divergence recovered by patching. Values above 1 (patching
overshoots and moves the output beyond the identity distribution) or
below 0 (patching moves the output \emph{away} from the identity
distribution, i.e., disrupts rather than mediates) are truncated for
layer ranking purposes.

We normalize by the per-sample baseline $D_i$ rather than using raw
patched JSD because the strength of personalization varies
substantially across samples: some triplets elicit strong
distributional shifts while others produce near-identical outputs.
Without normalization, layer importance scores would be dominated by
the few high-$D_i$ samples; normalizing yields a comparable fraction
in $[0, 1]$ that gives equal weight to every informative sample when
averaging across the dataset.

An NIE score near 1 indicates that, in this patching test, layer $\ell$'s
post-attention state at the last-token position is sufficient to move the
neutral run close to the cued next-token distribution. We therefore use
NIE to rank layers implicated in that output difference.

\subsection{Sample filter and layer ranking}
\label{sec:app-patching-filter}

Before averaging NIE across samples to rank layers, we exclude samples
with $D_i$ below the 40th percentile of the per-(model, condition)
distribution. This is not an arbitrary threshold but a methodological
prerequisite: when $D_i \approx 0$, the identity and neutral hidden
states are nearly identical at every layer, so patching one for the
other is uninformative---the causal intervention is only well-posed
when the two conditions produce detectably different outputs. Excluding
these samples removes noise from the layer ranking without biasing it.

The layer ranking is then $\ell_1, \ell_2, \ldots$ in descending
order of $\overline{\mathrm{NIE}(\ell)}$, averaged over the retained
samples.

\subsection{Cutoff choices across experiments}
\label{sec:app-patching-topk}

The same NIE ranking is used with two different cutoffs across the
paper:
\begin{itemize}\itemsep2pt
\item \textbf{Condition-specific top-10 layers} in \S\ref{sec:corr}.
$s_{\Delta A}$ is averaged across these ten layers before correlation,
which stabilizes the per-sample magnitude estimate.
\item \textbf{Pairwise union of the two dimensions' condition-level
top-10 sets} in \S\ref{sec:lin}. Depending on model and pair, this gives
12--26 unique layers on which the full raw contrasts are decomposed.
\item \textbf{Top-2 layers} in \S\ref{sec:abl} (direction ablation).
The projection hook is installed at the union of each dimensions' top two layers, two to six layers per dimension, to minimize intervention footprint, on
the reasoning that a smaller set of hooked layers is less likely to
disrupt unrelated computations (as evidenced by the MMLU results
reported in \S\ref{sec:abl}).
\end{itemize}



\subsection{Robustness of the ranking}
\label{sec:app-patching-robust}

The layer ranking is robust to the choice of divergence measure:
replacing JSD with the symmetric Kullback--Leibler divergence or the
total variation distance yields, in our tests, an overlap of 7--9 of
the top 10 layers across conditions. The final $L^\star$ chosen for
downstream use is thus not sensitive to reasonable perturbations of
the divergence definition, only to the underlying causal-mediation
logic.

\subsection{Computational resources.}
All experiments run on a single NVIDIA A100-SXM4 with 40\,GB of memory. The
five evaluated models are instruction-tuned checkpoints between 7B and 14B
parameters, loaded in bfloat16, which places an 8B model in roughly 16\,GB and
keeps the 14B models within the same device. The experiments reported here
comprise activation patching to identify $L^\star$ over nine single-cue
conditions on five models, cached-activation capture and per-sample
decomposition for three dimension pairs, the $\alpha$ sweep for the direction
ablation on three models, and the MMLU capability check, totalling
approximately 75 GPU-hours. No model is fine-tuned and no gradients are
computed, so all reported compute is inference. Dialogue construction and CAR
annotation use the GPT-4o API rather than local hardware.


\section{Behavioral Phenomenon and Significance Testing}
\label{sec:app-phenom}

\begin{figure}[t]
\centering
\includegraphics[width=\linewidth]{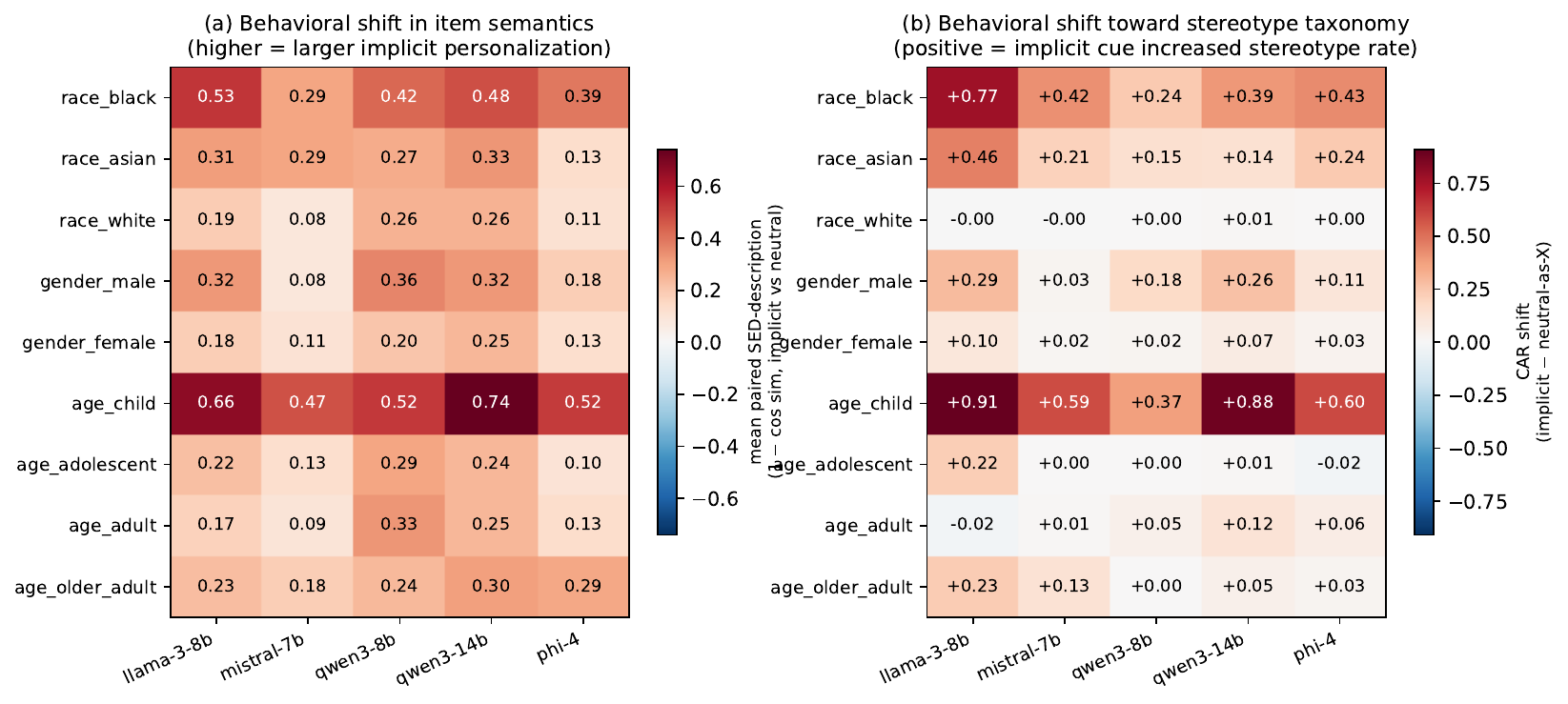}
\caption{Behavioral phenomenon at $K{=}5$ turns $\times$ movies across 5
models $\times$ 9 conditions. Left: mean paired SED-description per cell.
Right: mean CAR shift (implicit minus neutral scored as the same
identity). Both panels show effect sizes on a diverging colormap centered
at $0$---red a larger shift, blue none or anti-personalization---with no
significance read off the figure.}
\label{fig:phenom}
\end{figure}

Figure~\ref{fig:phenom} maps the behavioral phenomenon that
\S\ref{sec:corr} tests: the mean paired SED-description and the mean CAR
shift for every one of the five models across all nine single-dimension
conditions. It shows the effect-size landscape behind the significance counts
reported in the body---the two universal positive anchors (the Black-cue and
child-cue conditions), the universal null (the White-cue condition), and the
model-specific pattern elsewhere---and carries no significance markers, which
are reported per condition in \S\ref{sec:corr}.

\paragraph{Significance test behind the counts.}
A condition counts as showing a personalization shift when either behavioral
metric moves reliably away from neutral: a one-sided Welch $t$-test of the
identity-cued SED-description against the neutral split-half baseline
(\S\ref{sec:eval-behavior}), or a paired $t$-test of the CAR shift, clearing
an uncorrected $p<.05$. This is the test behind the per-model counts
reported in \S\ref{sec:corr} ($7/9$, $6/9$, $4/9$, $7/9$, $7/9$ for Llama,
Mistral, Qwen3-8B, Qwen3-14B, and Phi-4).

\section{Robustness of the Correlation Analysis}
\label{sec:app-corr-robust}
\label{sec:app-corr-concord}

This appendix tests the per-sample correlations of \S\ref{sec:corr}
against three concerns: multiple comparisons across the full grid, how the
two behavioral metrics relate, and reuse of the same samples for layer
selection and correlation.

Across the 35 cells scorable on both metrics, the $s_{\Delta A}$--SED and
$s_{\Delta A}$--CAR correlation columns themselves correlate at $r=.70$
($p=2.75\times10^{-6}$). Conditions in which activation magnitude tracks
broad semantic movement therefore also tend to be those in which it tracks
identity-associated content: the Black- and child-cue anchors are positive
on both metrics across all five models.

The correspondence is not exact because the metrics measure different kinds
of change. SED registers any thematic movement, whereas CAR registers only
movement along a fixed identity taxonomy. The clearest divergence is
Mistral's male-cue condition (SED $r=+.37$; CAR $r=-.16$): activation
magnitude tracks a semantic change there, but not increased alignment with
the male-associated taxonomy. We therefore interpret convergence on the
strong conditions as complementary evidence and the remaining differences
as information about the kind of behavioral change, rather than requiring
the two metrics to agree cell by cell.




\section{Intersectional Interaction: Definition and Full Results}
\label{sec:app-interaction}

This appendix collects the interaction material deferred from
\S\ref{sec:lin}: the definition and testing protocol of the interaction
contrast $\psi$ with its complete per-model results, the cell-level view
behind the body's showcase example, the decomposition-weight and
sample-specificity summary, and the behavioral sub-additivity
quantification with its two negative controls.
All results are reported as tables computed from the canonical runs.

\subsection{The interaction contrast $\psi$ and its full results}
\label{sec:app-int-def}

The reshaping test of \S\ref{sec:lin} uses the two-way interaction
contrast of the $2{\times}2$ factorial, a difference of differences.
Writing $a_i, a_j$ for two levels of dimension~A, $b_i, b_j$ for two
levels of dimension~B, and $\bar y_{ab}$ for the cell mean of a
neutral-normalized CAR outcome,
\begin{equation}
\psi \;=\;
  (\bar{y}_{a_i b_i} - \bar{y}_{a_j b_i})
  \;-\;
  (\bar{y}_{a_i b_j} - \bar{y}_{a_j b_j}) :
\end{equation}
the A-gap measured under level $b_i$ of~B, minus the same A-gap under
$b_j$. A non-zero $\psi$ says the size of one dimension's expression
depends on the level of the other. Throughout the intersectional
analysis, CAR$_{\rm dim}$ counts items tagged with \emph{any}
non-neutral level of that dimension (CAR$_{\rm race}$ counts Asian-,
Black-, and White-tagged items alike), so cell means and contrasts are
well-defined even for levels whose own lexicon is sparse; this differs
from the single-dimension CAR of \S\ref{sec:corr}, which scores each
condition against its own level's taxonomy. The contrast is robust to base-rate
differences between groups: because CAR is neutral-normalized, each
group's standing tag rate is already subtracted, and because $\psi$ is a
difference of differences, any residual tendency for one group to carry
more tagged items enters both bracketed terms and cancels. We compute
$\psi$ on the $n=50$ mixed cells for every level pairing of every
dimension pair and every CAR outcome, and test it with a paired $t$-test
under Benjamini--Hochberg correction within each (model, pair, DV)
family.

Table~\ref{tab:app-psi-sig} lists every contrast that survives
correction: $12$ of the $405$ tested. The significant structure is
concentrated on three models. Llama carries five (three on
race$\times$gender, the pattern unpacked in \S\ref{sec:lin}, and two on
gender$\times$age, both involving the older-adult level); Qwen3-14B
carries six, all on race$\times$age and all involving the adolescent
level; Phi-4 carries one. Mistral and Qwen3-8B carry none. The
interaction is therefore real but sparse and model-specific, not a
universal effect.

\begin{table*}[t]
\centering \footnotesize
\begin{tabular}{ll l l r r r}
\toprule
Model & Pair & DV & Contrast & $\psi$ & $d$ & $q_{\rm BH}$ \\
\midrule
llama-3-8b & race$\times$gender & CAR$_{\rm race}$ & (Black$-$White)$\times$(Female$-$Male) & $+0.168$ & $+0.40$ & $0.021$ \\
 & race$\times$gender & CAR$_{\rm gender}$ & (Asian$-$Black)$\times$(Female$-$Male) & $+0.144$ & $+0.44$ & $0.010$ \\
 & race$\times$gender & CAR$_{\rm race+gender}$ & (Asian$-$White)$\times$(Female$-$Male) & $+0.072$ & $+0.41$ & $0.016$ \\
 & gender$\times$age & CAR$_{\rm gender}$ & (Female$-$Male)$\times$(Child$-$Older) & $-0.156$ & $-0.41$ & $0.026$ \\
 & gender$\times$age & CAR$_{\rm gender}$ & (Female$-$Male)$\times$(Adol.$-$Older) & $-0.180$ & $-0.39$ & $0.026$ \\
\addlinespace[2pt]
qwen3-14b & race$\times$age & CAR$_{\rm race}$ & (Black$-$White)$\times$(Adol.$-$Older) & $+0.127$ & $+0.56$ & $0.009$ \\
 & race$\times$age & CAR$_{\rm race}$ & (Black$-$White)$\times$(Adol.$-$Adult) & $+0.107$ & $+0.53$ & $0.009$ \\
 & race$\times$age & CAR$_{\rm race}$ & (Black$-$White)$\times$(Adol.$-$Child) & $+0.140$ & $+0.48$ & $0.018$ \\
 & race$\times$age & CAR$_{\rm race}$ & (Asian$-$Black)$\times$(Adol.$-$Older) & $-0.107$ & $-0.41$ & $0.036$ \\
 & race$\times$age & CAR$_{\rm race+age}$ & (Black$-$White)$\times$(Adol.$-$Older) & $+0.045$ & $+0.48$ & $0.026$ \\
 & race$\times$age & CAR$_{\rm race+age}$ & (Black$-$White)$\times$(Adol.$-$Adult) & $+0.044$ & $+0.47$ & $0.026$ \\
\addlinespace[2pt]
phi-4 & gender$\times$age & CAR$_{\rm age}$ & (Female$-$Male)$\times$(Adol.$-$Older) & $-0.150$ & $-0.56$ & $0.003$ \\
\bottomrule
\end{tabular}

\caption{All BH-significant interaction contrasts across the canonical
runs: $12$ of $405$ tested (five models $\times$ three pairs $\times$
three CAR outcomes, all level pairings). $\psi$ is the
difference-of-differences on the neutral-normalized outcome, $d$ is
Cohen's $d$, and $q_{\rm BH}$ is the corrected $p$ within the (model,
pair, DV) family.}
\label{tab:app-psi-sig}
\end{table*}

\subsection{Cell-level view of the focus pair}
\label{sec:app-int-cells}

Table~\ref{tab:app-psi-cells} gives the cell means behind the two
headline Llama race$\times$gender contrasts. Race expression appears
only under a female cue and peaks in the Black-female cell ($+0.216$
over neutral, against $+0.004$ or below in every male-cued cell), while
gender expression is robust in every cell ($+0.08$ to $+0.14$)
\emph{except} Black-female, where it collapses to $+0.028$: the pattern
summarized in \S\ref{sec:lin} as race dominating that intersection at
the gender cue's expense. Consistent with a fixed-length recommendation
list imposing a content budget, the two tag rates anti-correlate per
sample in five of the six mixed cells ($r=-.28$ to $-.49$); the
exception is Asian-female ($r=+.06$), the one cell that co-expresses
both cues, where items can carry both tags at once. Its co-tagging is
also what drives the third significant contrast, on
CAR$_{\rm race+gender}$.

\begin{table}[t]
\centering \small
\begin{tabular}{l rr}
\toprule
Mixed cell & CAR$_{\rm race}$ & CAR$_{\rm gender}$ \\
\midrule
Asian $\times$ female & $+0.120$ & $+0.144$ \\
Asian $\times$ male & $-0.012$ & $+0.112$ \\
Black $\times$ female & $+0.216$ & $+0.028$ \\
Black $\times$ male & $+0.004$ & $+0.140$ \\
White $\times$ female & $-0.048$ & $+0.080$ \\
White $\times$ male & $-0.092$ & $+0.116$ \\
\bottomrule
\end{tabular}

\caption{Neutral-normalized CAR$_{\rm race}$ and CAR$_{\rm gender}$ cell
means for the six Llama race$\times$gender mixed cells ($n=50$ each).
Race expression is amplified and gender expression compressed in the
Black-female cell; Asian-female co-expresses both.}
\label{tab:app-psi-cells}
\end{table}

\subsection{Decomposition weights and sample specificity}
\label{sec:app-int-null}

Across all $15$ model$\times$pair combinations the mean fitted
weights fall below the additive target $(1,1)$.

The informative control substitutes a \emph{different} same-cell
sample's $(v_A,v_B)$ basis. This lowers fit from $.576$--$.704$ to
$.038$--$.156$ (overall mean $.095$), close to the reference obtained
from another mixed sample alone. The own basis wins in all 15
model--pair combinations, by $.522$--$.577$ (two-sided sign test
$p<10^{-4}$). The decomposition is therefore sample-specific, not a
generic cell-level geometry. The residual fit leaves roughly 30--42\% of
$\lVert v_{AB}\rVert$ unexplained, so the claim the data support is a
substantial, sample-specific linear component rather than exact addition.

A second control asks whether one cue alone is sufficient. Refitting each
mixed vector with $v_A$ or $v_B$ alone gives an overall best-one mean of
$.443$, versus $.646$ for both terms. The two-term model improves on the
better single term in every one of 6,185 records, with model--pair mean
gains of $.186$--$.226$ (Table~\ref{tab:oneterm}). The mixed contrast
therefore contains information associated with both single-cue
contrasts, rather than one cue plus a redundant term.

\begin{table*}[t]
\centering \small
\setlength{\tabcolsep}{4pt}
\begin{tabular}{ll rrrrr}
\toprule
Model & Pair & $\bar\rho_{\rm two}$ & $\bar\rho_A$ & $\bar\rho_B$ & $\bar\rho_{\rm best\,one}$ & $\bar\rho_{\rm two}-\bar\rho_{\rm best\,one}$ \\
\midrule
llama-3-8b & race$\times$gender & $0.699$ & $0.373$ & $0.292$ & $0.486$ & $0.213$ \\
 & race$\times$age & $0.704$ & $0.374$ & $0.292$ & $0.478$ & $0.226$ \\
 & gender$\times$age & $0.695$ & $0.342$ & $0.342$ & $0.490$ & $0.206$ \\
\addlinespace[2pt]
mistral-7b & race$\times$gender & $0.676$ & $0.391$ & $0.394$ & $0.487$ & $0.188$ \\
 & race$\times$age & $0.673$ & $0.395$ & $0.374$ & $0.487$ & $0.186$ \\
 & gender$\times$age & $0.665$ & $0.364$ & $0.360$ & $0.468$ & $0.197$ \\
\addlinespace[2pt]
qwen3-8b & race$\times$gender & $0.636$ & $0.320$ & $0.330$ & $0.422$ & $0.213$ \\
 & race$\times$age & $0.614$ & $0.320$ & $0.324$ & $0.407$ & $0.206$ \\
 & gender$\times$age & $0.615$ & $0.324$ & $0.326$ & $0.413$ & $0.202$ \\
\addlinespace[2pt]
qwen3-14b & race$\times$gender & $0.577$ & $0.307$ & $0.311$ & $0.387$ & $0.191$ \\
 & race$\times$age & $0.578$ & $0.306$ & $0.315$ & $0.381$ & $0.196$ \\
 & gender$\times$age & $0.576$ & $0.297$ & $0.307$ & $0.379$ & $0.197$ \\
\addlinespace[2pt]
phi-4 & race$\times$gender & $0.670$ & $0.362$ & $0.364$ & $0.462$ & $0.208$ \\
 & race$\times$age & $0.658$ & $0.364$ & $0.347$ & $0.450$ & $0.208$ \\
 & gender$\times$age & $0.661$ & $0.352$ & $0.353$ & $0.450$ & $0.211$ \\
\bottomrule
\end{tabular}

\caption{One-term vs.\ two-term decomposition fit for all $15$
model$\times$pair combinations. $\bar\rho_{\rm two}$ is the own-sample
two-basis fit reported above; $\bar\rho_{A}$ and
$\bar\rho_{B}$ fit $v_{AB}$ with a single single-cue direction. Both
one-term fits fall far short of the two-term fit, so neither cue alone
explains the mixed-cue vector.}
\label{tab:oneterm}
\end{table*}

\subsection{Behavioral sub-additivity and two negative controls}
\label{sec:app-int-joint}

We quantify the behavioral sub-additivity summarized in
\S\ref{sec:lin} as follows. For each of the three ablation models we
compute the paired SED-description against neutral for the mixed
response and its two \emph{dose-matched} single-cue responses, where
each pure context carries exactly the cues it contributes to the mixed
one, so that under strict additivity the observed mixed shift would
equal $\mathrm{SED}_A + \mathrm{SED}_B$. Regressing observed on
predicted, the fitted slope is $0.39$--$0.49$ across the three models,
$68$--$85\%$ of samples fall below the additive diagonal, and a one-sided
$t$-test on the per-sample residuals is decisive ($p<10^{-37}$ on every
model). The mean relative shortfall of
$-28\%$ to $-38\%$ says, in plain terms, that two co-present cues
deliver only about two-thirds of what independently adding their
single-cue shifts would predict.

Two negative results bound this composition from above, and we record
them so neither is mistaken for its opposite. First, \emph{joint marking
does not exceed the marginals}. If mixed contexts bound the two
dimensions into jointly stereotyped items, the co-tag rate
CAR$_{\rm both}$ would exceed the independence prediction
CAR$_A \times$ CAR$_B$. It does not: observed co-tagging falls
\emph{below} the independence product in $86$ of $130$ cells (sign test
$p = 6.9\times10^{-6}$), and the only panels that consistently exceed
independence, Llama and Phi-4 race$\times$age, are single-title
artifacts. Second, the strictest additivity test on the pure
conditions, $\Delta_{\rm Int} = M - P_A - P_B + N$ (super-additivity),
is nearly null. On the first dimension's tag rate, $0$ of $90$
arm-level contrasts survive within-family BH (mean
$\Delta_{\rm Int} = -0.0007$); on the second dimension's, $3$ of $90$
survive, all on Mistral gender$\times$age's age outcome, one of which
survives global correction. This near-null is a different interaction
notion from $\psi$: a design can be perfectly additive over its pure
conditions and still show a non-zero $\psi$, so the near-null
$\Delta_{\rm Int}$ is compatible with the significant $\psi$ contrasts
rather than in tension with them. Together the two results bound the
interaction from above, almost no amplification and no super-marginal
joint marking, while leaving intact the sparse, sign-consistent $\psi$
modulation of Table~\ref{tab:app-psi-sig}.
\section{Full Ablation Grids}
\label{sec:app-ablation-grids}

\S\ref{sec:abl} shows the SED-description ablation grid for Mistral-7B
and Qwen3-8B and reports the CAR effect only for the
strong-personalization cells; this appendix completes both grids with
the remaining Llama-3-8B SED block and the full three-model CAR grid,
and additionally reports every cell at exact projection ($\alpha=1$).

\paragraph{SED-description grid.}
Table~\ref{tab:ablation-full} is the Llama-3-8B block of the
SED-description grid at each cell's $\alpha^\star$; together with
Table~\ref{tab:ablation} in the body it covers all $18$ target cells
(three models $\times$ three pairs $\times$ two targets). Across the
$18$ cells, ablation suppresses the predicted shift in $15$ and
matches or outperforms the prompt in $16$. The three non-suppressing
cells, Llama's two gender targets and Mistral's age target on
gender$\times$age, are all near-zero.

\begin{table*}[t]
\centering \small
\begin{tabular}{ll l c r r r r}
\toprule
Model & Pair & target & $\alpha^\star$ & $\Delta_{\rm pure(t)}$ & $d$ & Prompt~$\Delta$ & Ratio \\
\midrule
\multirow{6}{*}{\textbf{llama-3-8b}}
 & \multirow{2}{*}{race\,$\times$\,gender} & race   & 5.0 & $+0.086^{***}$ & $+0.56$ & $+0.006$ & \textbf{13.5$\times$} \\
 &                                         & gender & 4.0 & $+0.005$       & $+0.05$ & $+0.004$ & \textbf{1.2$\times$} \\
\addlinespace[1pt]
 & \multirow{2}{*}{race\,$\times$\,age}    & race   & 5.0 & $+0.078^{***}$ & $+0.54$ & $+0.016$ & \textbf{5.0$\times$} \\
 &                                         & age    & 5.0 & $+0.022^{***}$ & $+0.19$ & $+0.016$ & \textbf{1.4$\times$} \\
\addlinespace[1pt]
 & \multirow{2}{*}{gender\,$\times$\,age}  & gender & 4.0 & $+0.003$       & $+0.03$ & $+0.017$ & \textbf{0.2$\times$} \\
 &                                         & age    & 5.0 & $+0.038^{***}$ & $+0.33$ & $+0.009$ & \textbf{4.5$\times$} \\
\bottomrule
\end{tabular}

\caption{Llama-3-8B block of the direction-ablation grid at each cell's
$\alpha^\star$, measured on SED-description; the Mistral-7B and Qwen3-8B
blocks are Table~\ref{tab:ablation} in the body, reporting the same
quantities.
$\Delta_{\mathrm{pure}(t)}$ is the per-cell shift away from the
target-dimension reference response and $d$ is Cohen's $d$. Ratio is
ablation over prompt. Stars: $^*\,p<.05$, $^{**}\,p<.01$,
$^{***}\,p<.001$.}
\label{tab:ablation-full}
\end{table*}

\paragraph{CAR grid, both sides.}
Table~\ref{tab:app-car} gives all $18$ rows of the target-taxonomy CAR
effect at \texttt{vk=dim}, each at its own best $\alpha^\star$, with
bootstrap $95\%$ CIs and the matched ``ignore-demographic'' prompt
baseline. The headline negative cells are Mistral race$\times$gender on
the gender target ($\Delta\mathrm{CAR}=-0.169$, $p=1.6\times10^{-28}$,
the largest reduction in the grid), Qwen3-8B race$\times$age ($-0.130$,
$p=3.8\times10^{-78}$) and race$\times$gender ($-0.092$,
$p=1.9\times10^{-21}$) on the race target, Mistral and Qwen3-8B
gender$\times$age on the gender target ($-0.039$, $p=1.3\times10^{-6}$;
$-0.030$, $p=1.8\times10^{-3}$), and Llama race$\times$gender on the
race target ($-0.030$, $p=.02$).

Two properties of the grid are worth stating. First, the effect
concentrates on race and gender targets: every \emph{age}-target row is
null or small---the one significant age reduction is Mistral
gender$\times$age at $-0.024$ ($p=2.0\times10^{-4}$)---consistent with
age being the least-personalized dimension in \S\ref{sec:corr}. Second,
on Mistral the two metrics dissociate \emph{by dimension}: its
race-target rows are null on CAR ($+0.006$, $+0.001$) despite carrying
some of the grid's largest semantic shifts on SED-description
($d=1.22$--$1.59$, Table~\ref{tab:ablation}), while its
gender-target ablations register strongly on both metrics. For the SED-versus-CAR contrast, Mistral
changing themes without flipping taxonomy labels, is therefore a
property of its race expression specifically, not of the model as a whole.

\begin{table*}[t]
\centering \small
\begin{tabular}{ll l c r c r}
\toprule
Model & Pair & target & $\alpha^\star$ & $\Delta$CAR$_{(t)}$ & 95\% CI & Prompt~$\Delta$ \\
\midrule
\multirow{6}{*}{\textbf{llama-3-8b}}
 & \multirow{2}{*}{race\,$\times$\,gender} & race & 2.0 & $-0.030^{*}$ & $[-0.057, -0.006]$ & $+0.025$ \\
 &  & gender & 0.5 & $+0.014$ & $[+0.006, +0.022]$ & $-0.076$ \\
\addlinespace[1pt]
 & \multirow{2}{*}{race\,$\times$\,age} & race & 0.5 & $-0.009$ & $[-0.020, +0.003]$ & $-0.004$ \\
 &  & age & 0.5 & $+0.018$ & $[+0.007, +0.030]$ & $-0.073$ \\
\addlinespace[1pt]
 & \multirow{2}{*}{gender\,$\times$\,age} & gender & 0.5 & $+0.006$ & $[-0.003, +0.015]$ & $-0.066$ \\
 &  & age & 0.5 & $+0.002$ & $[-0.012, +0.015]$ & $-0.054$ \\
\midrule
\multirow{6}{*}{\textbf{mistral-7b}}
 & \multirow{2}{*}{race\,$\times$\,gender} & race & 1.0 & $+0.006$ & $[+0.001, +0.011]$ & $+0.005$ \\
 &  & gender & 5.0 & $\mathbf{-0.169}^{***}$ & $[-0.193, -0.146]$ & $+0.022$ \\
\addlinespace[1pt]
 & \multirow{2}{*}{race\,$\times$\,age} & race & 0.5 & $+0.001$ & $[-0.003, +0.004]$ & $+0.011$ \\
 &  & age & 0.5 & $-0.002$ & $[-0.008, +0.003]$ & $-0.012$ \\
\addlinespace[1pt]
 & \multirow{2}{*}{gender\,$\times$\,age} & gender & 5.0 & $\mathbf{-0.039}^{***}$ & $[-0.055, -0.022]$ & $+0.053$ \\
 &  & age & 5.0 & $-0.024^{***}$ & $[-0.038, -0.011]$ & $-0.024$ \\
\midrule
\multirow{6}{*}{\textbf{qwen3-8b}}
 & \multirow{2}{*}{race\,$\times$\,gender} & race & 5.0 & $\mathbf{-0.092}^{***}$ & $[-0.109, -0.076]$ & $-0.044$ \\
 &  & gender & 1.0 & $-0.002$ & $[-0.012, +0.007]$ & $+0.011$ \\
\addlinespace[1pt]
 & \multirow{2}{*}{race\,$\times$\,age} & race & 4.0 & $\mathbf{-0.130}^{***}$ & $[-0.140, -0.121]$ & $-0.049$ \\
 &  & age & 1.0 & $+0.009$ & $[-0.002, +0.019]$ & $+0.040$ \\
\addlinespace[1pt]
 & \multirow{2}{*}{gender\,$\times$\,age} & gender & 5.0 & $-0.030^{**}$ & $[-0.048, -0.012]$ & $+0.007$ \\
 &  & age & 5.0 & $-0.013$ & $[-0.035, +0.008]$ & $+0.019$ \\
\bottomrule
\end{tabular}

\caption{Full CAR ablation grid at \texttt{vk=dim}, both ablation sides
per pair (rows labeled by the ablated \emph{target} dimension), each
cell at its own $\alpha^\star$. $\Delta$CAR$_{(t)}$ is the change in the
target dimension's taxonomy CAR after ablation (negative $=$
stereotype-labeled items removed); stars are from the one-sided Wilcoxon
test on the target dimension's per-sample $\Delta$CAR ($^{*}p<.05$,
$^{**}p<.01$, $^{***}p<.001$); brackets are bootstrap $95\%$ CIs.
``Prompt~$\Delta$'' is the matched ``ignore-demographic'' baseline on
the same target. Bold marks the largest significant CAR reductions.}
\label{tab:app-car}
\end{table*}

\paragraph{Exact projection ($\alpha=1$).}
The grids above report each cell at its best sweep strength
$\alpha^\star$. For completeness, Table~\ref{tab:alpha1} gives every cell
at $\alpha=1$, where the hook deletes the component along
$\hat v_{\mathrm{dim}}$ exactly (Eq.~\eqref{eq:ablation}). The effect is
present at exact projection and grows with $\alpha$: $11$ of the $18$ SED
targets shift significantly in the predicted direction at $\alpha=1$, and
the strongest CAR cells already register, with Qwen3-8B's race target
falling by $0.059$ on race$\times$age and $0.019$ on race$\times$gender.

\begin{table*}[t]
\centering \small
\begin{tabular}{ll l r r r}
\toprule
Model & Pair & target & $\Delta_{\rm pure(t)}$ & $d$ & $\Delta$CAR$_{(t)}$ \\
\midrule
\multirow{6}{*}{\textbf{llama-3-8b}}
 & \multirow{2}{*}{race\,$\times$\,gender} & race & $+0.018^{**}$ & $+0.15$ & $-0.017$ \\
 &  & gender & $-0.002$ & $-0.03$ & $+0.027$ \\
\addlinespace[1pt]
 & \multirow{2}{*}{race\,$\times$\,age} & race & $+0.018^{***}$ & $+0.17$ & $+0.008$ \\
 &  & age & $+0.015^{***}$ & $+0.15$ & $+0.026$ \\
\addlinespace[1pt]
 & \multirow{2}{*}{gender\,$\times$\,age} & gender & $-0.002$ & $-0.03$ & $+0.006$ \\
 &  & age & $+0.010^{*}$ & $+0.11$ & $+0.016$ \\
\midrule
\multirow{6}{*}{\textbf{mistral-7b}}
 & \multirow{2}{*}{race\,$\times$\,gender} & race & $+0.025^{***}$ & $+0.24$ & $+0.006$ \\
 &  & gender & $+0.028^{***}$ & $+0.31$ & $-0.009$ \\
\addlinespace[1pt]
 & \multirow{2}{*}{race\,$\times$\,age} & race & $+0.026^{***}$ & $+0.26$ & $+0.001$ \\
 &  & age & $+0.007^{*}$ & $+0.08$ & $-0.002$ \\
\addlinespace[1pt]
 & \multirow{2}{*}{gender\,$\times$\,age} & gender & $+0.000$ & $+0.01$ & $-0.005$ \\
 &  & age & $-0.011$ & $-0.15$ & $-0.011^{**}$ \\
\midrule
\multirow{6}{*}{\textbf{qwen3-8b}}
 & \multirow{2}{*}{race\,$\times$\,gender} & race & $+0.005$ & $+0.04$ & $-0.019^{***}$ \\
 &  & gender & $+0.010^{*}$ & $+0.10$ & $-0.002$ \\
\addlinespace[1pt]
 & \multirow{2}{*}{race\,$\times$\,age} & race & $+0.079^{***}$ & $+0.33$ & $-0.059^{***}$ \\
 &  & age & $+0.022^{***}$ & $+0.15$ & $+0.009$ \\
\addlinespace[1pt]
 & \multirow{2}{*}{gender\,$\times$\,age} & gender & $+0.008$ & $+0.07$ & $+0.018$ \\
 &  & age & $+0.011$ & $+0.07$ & $+0.003$ \\
\bottomrule
\end{tabular}

\caption{All $18$ ablation cells at exact projection ($\alpha=1$),
\texttt{vk=dim}: the SED-description shift away from the target
reference $\Delta_{\mathrm{pure}(t)}$ with Cohen's $d$, and the
target-taxonomy $\Delta$CAR$_{(t)}$. Stars: $^{*}p<.05$, $^{**}p<.01$,
$^{***}p<.001$, from the same tests as Tables~\ref{tab:ablation}
and~\ref{tab:app-car} (SED: one-sided $t$, predicted positive
direction; CAR: one-sided Wilcoxon, predicted negative direction);
stars are shown only for movements in the predicted direction.}
\label{tab:alpha1}
\end{table*}

\paragraph{Projection-magnitude-matched sham.}
\label{sec:app-sham-matched}
The original random-direction reference matches the direction vector's norm,
but the hook normalizes that vector before projection. We therefore rerun a
stricter sham whose random-direction perturbation matches the real
per-token magnitude $\alpha|h^\top\hat v|$. This control is generated
end-to-end as its own run, so its ablation arm differs from
Table~\ref{tab:alpha1} in the third decimal. At exact projection the matched
sham moves SED-description by $+0.015$ against the ablation's $+0.079$ on
Qwen3-8B race$\times$age, and the ablation's target-CAR reductions
($-0.060$ and $-0.018$ on the two Qwen3-8B race cells) do not appear under
it ($+0.009$ and $+0.005$). At $\alpha=5$ the
sham produces nonzero semantic drift, while the Qwen3-8B target-CAR
reductions remain specific to the learned direction.
The large-strength semantic effect should therefore be read as a mixture of
direction-specific removal and generic perturbation, while the exact-projection
result supplies the cleaner causal test.

\section{MMLU Evaluation and Vector-Source Robustness}
\label{sec:app-mmlu}

This appendix documents two supporting details for the direction
ablation of \S\ref{sec:abl}: the protocol behind the MMLU
capability-preservation check and a
robustness comparison showing that the ablation's effect does not
depend on the specific vector source used to define the ablated
direction.

\subsection{MMLU protocol}
\label{sec:app-mmlu-protocol}

The capability check evaluates each of the three ablation models
(Llama-3-8B, Mistral-7B, Qwen3-8B) on five MMLU subjects---elementary
mathematics, college biology, high-school world history, philosophy,
and sociology---at $100$ questions each, for $500$ questions per
condition, posed in the standard few-shot MMLU format; accuracy is the
macro-average over the five per-subject accuracies.
We evaluate no intervention (\emph{baseline}) and the direction ablation
at $\alpha \in \{1, 3, 5\}$. Baseline macro-accuracy is $0.704$ on
Llama-3-8B, $0.616$ on Mistral-7B and $0.634$ on Qwen3-8B. At $\alpha=1$
the largest change is $-0.006$ (Qwen3-8B); at $\alpha=5$ it is $-0.028$,
again on Qwen3-8B, while Llama and Mistral stay within $0.006$ of baseline
at every strength. For every ablation
condition the intervention is \emph{active during answer
generation}---the same forward pre-hook, at the same $L^\star$ hook
layers, that produces the behavioral
effect---so the check measures
the intervention's true collateral cost on general reasoning.
Llama-3-8B is evaluated with the few-shot block as a plain
continuation rather than wrapped in its chat template, whose turn
structure breaks the benchmark's exemplars-then-question format and
collapses the baseline to near chance; under the plain format its
baseline is $0.704$.

The explicit ``ignore-demographic'' prompt of \S\ref{sec:abl} is
excluded from this comparison because its effect on Qwen3-8B is
anomalous: it \emph{raises} macro-accuracy from $0.634$ to $0.734$,
nearly uniformly across all five subjects ($+0.03$ to $+0.17$). A
content-irrelevant instruction about demographics has no mechanism to
improve elementary mathematics and philosophy alike; the pattern is
instead consistent with a prompt-format artifact, where adding a
system prompt changes the chat-template structure that Qwen3's answer
formatting is sensitive to, and the other two models move negligibly
under the same prompt (Llama $+0.002$, Mistral $-0.004$). We
therefore do not read the prompt condition as evidence of a
capability change in either direction.

\subsection{Vector-source robustness}
\label{sec:app-vectorsource}

\paragraph{Three ways to define the ablated direction.}
The ablation of \S\ref{sec:abl} projects out the unit direction of the
\emph{pure single-dimension contrast} (\texttt{vk=dim}): for the
ablated dimension $A$, $v_A = \mathbf{A}(C_{\mathrm{pure}\text{-}A}) -
\mathbf{A}(C_{\mathrm{neu}})$ at $L^\star$, the dose-matched
raw single-cue activation contrast used throughout the paper
(\S\ref{sec:eval-deltaa}). Two alternative estimators of ``the $A$
direction'' are available. \texttt{vk=intersect} takes the
mixed-versus-dose-matched-pure contrast
($\mathbf{A}(C_{\mathrm{mixed}}) -
\mathbf{A}(C_{\mathrm{pure}\text{-}B})$), so it isolates $A$'s
contribution from \emph{within} a two-cue context and thus carries
additional intersectional signal. \texttt{vk=explicit} takes the
contrast between an explicitly-stated-demographic context and the
neutral context. All three name the same nominal direction by a
different activation difference; \S\ref{sec:abl} uses \texttt{dim}
because it is the simplest such difference, is the most aligned with
the activation-steering prior art, and---unlike intersect---
requires no dependence on the mixed-context data.

\paragraph{The effect is qualitatively source-invariant
(Table~\ref{tab:app-vk}).}
Table~\ref{tab:app-vk} reruns the ablation under each of the three
vector sources: three models $\times$ three pairs, $\mathrm{ablate}_A$
on description-SED, each source at its best sweep strength. The
qualitative picture of \S\ref{sec:abl} reproduces under all three: on
the strong-personalization cells the target-dimension shift is positive
and Wilcoxon-significant in the predicted direction under every source,
and the underlying sweeps rise monotonically with $\alpha$ while the original
random-direction reference stays near zero. The central claim that projecting out the
cue-induced dimension direction suppresses the target-dimension response is
therefore not an artifact of the particular contrast chosen to define
that direction.

\begin{table*}[t]
\centering \small
\begin{tabular}{ll rrr}
\toprule
Model & Pair & \texttt{dim} & \texttt{intersect} & \texttt{explicit} \\
\midrule
llama-3-8b & race$\times$gender & $+0.086^{***}$ & $+0.078^{***}$ & $+0.056^{***}$ \\
 & race$\times$age & $+0.078^{***}$ & $+0.062^{***}$ & $+0.041^{***}$ \\
 & gender$\times$age & $+0.003$ & $+0.006$ & $+0.016^{**}$ \\
\addlinespace[2pt]
mistral-7b & race$\times$gender & $+0.226^{***}$ & $+0.322^{***}$ & $+0.152^{***}$ \\
 & race$\times$age & $+0.154^{***}$ & $+0.375^{***}$ & $+0.022^{***}$ \\
 & gender$\times$age & $+0.028^{***}$ & $+0.025^{***}$ & $+0.004^{*}$ \\
\addlinespace[2pt]
qwen3-8b & race$\times$gender & $+0.176^{***}$ & $+0.103^{***}$ & $+0.118^{***}$ \\
 & race$\times$age & $+0.306^{***}$ & $+0.113^{***}$ & $+0.053^{***}$ \\
 & gender$\times$age & $+0.078^{***}$ & $+0.150^{***}$ & $+0.027^{***}$ \\
\bottomrule
\end{tabular}

\caption{Vector-source robustness: the $\mathrm{ablate}_A$
target-dimension shift on description-SED at each source's best sweep
strength, per (model, pair). Stars: one-sided Wilcoxon in the predicted
direction, $^{*}p<.05$, $^{**}p<.01$, $^{***}p<.001$. The
\texttt{dim} column matches the corresponding rows of
Tables~\ref{tab:ablation} and~\ref{tab:ablation-full}.}
\label{tab:app-vk}
\end{table*}

\paragraph{What differs across sources.}
The three sources are not identical in magnitude or in the relative
ordering of the two dimensions within a pair, and we do not claim they
are. \texttt{intersect} produces the largest race effects on Mistral,
where it drives the race target substantially higher than \texttt{dim}
on both race$\times$gender and race$\times$age, consistent with its
carrying extra mixed-context signal, though on the other two models
\texttt{dim} is the stronger race-target source;
\texttt{explicit} is the most variable across cells and is the
weakest on some, falling to near the sham for the race target on
Mistral race$\times$age. It can also move a target the primary source
does not: on Qwen3-8B race$\times$age the age target's CAR falls by
$0.155$ at $\alpha=5$ under \texttt{explicit} but shows no reduction
under \texttt{dim}, the clearest single case of source dependence. The ceiling-of-sweep behavior ($\alpha^\star$
at or near $5$) is shared by all three; the stricter matched-perturbation sham
is analyzed in Appendix~\ref{sec:app-sham-matched}. We
adopt \texttt{dim} as the primary source not because it is uniformly
the strongest---it is not---but because it is the simplest and most
prior-art-aligned estimator and introduces no methodological
dependence on the intersectional data; the two alternatives serve to
corroborate, rather than to change, the conclusion of \S\ref{sec:abl}.

\begin{figure}[t]
\centering
\includegraphics[width=\linewidth]{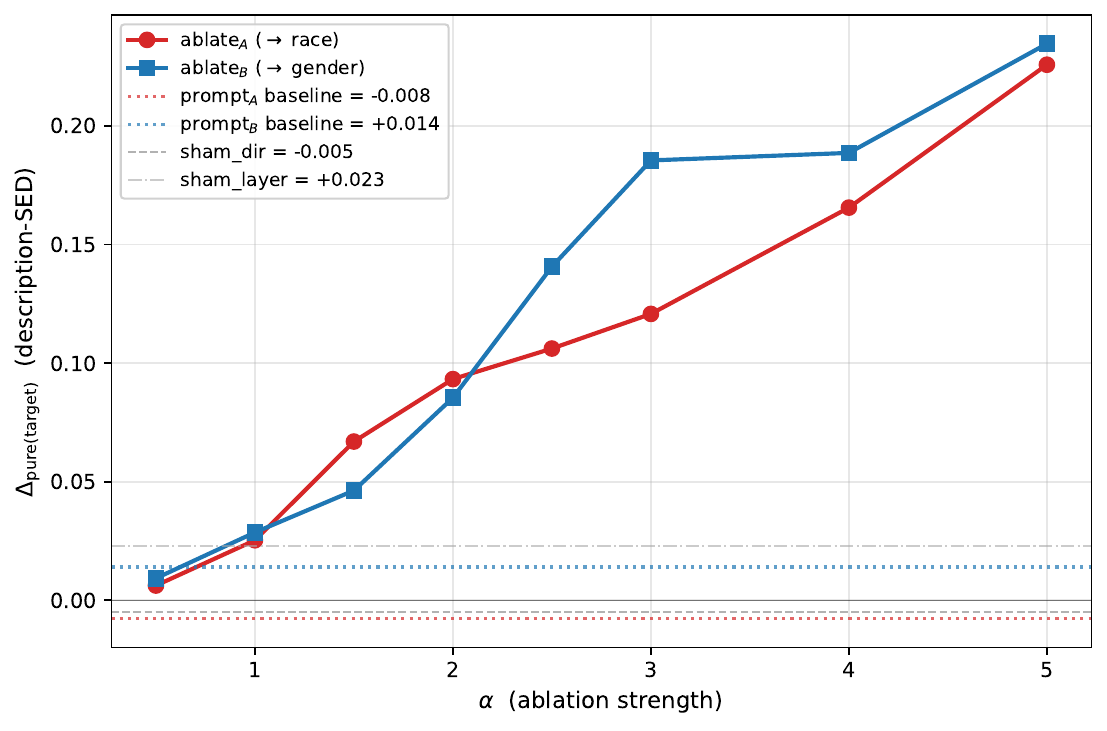}
\caption{Direction ablation vs.\ prompt vs.\ sham on Mistral-7B
race$\times$gender, SED-description. Both ablation targets
($\mathrm{ablate}_A\!\to\!$race, $\mathrm{ablate}_B\!\to\!$gender) rise
monotonically with $\alpha$; the two ``ignore-demographic'' prompt baselines
are flat and near zero, the race-side one slightly negative (that prompt
moves the output toward the cue), and the original direction- and layer-shams
hug zero.}
\label{fig:backfire}
\end{figure}

\paragraph{The prompt can backfire where the ablation does not.}
Figure~\ref{fig:backfire} isolates the clearest case from the
\texttt{vk=dim} grid, in SED-description terms: on Mistral race$\times$gender
both ablation targets climb monotonically with $\alpha$, while the two
explicit-prompt baselines sit flat near zero. The race-side prompt is
slightly negative ($-0.008$), so telling the model to ignore race moves
its recommendations \emph{toward} the stereotype; the gender-side prompt
is only marginally positive ($+0.014$). The sham references hug zero. This is the SED-description
view of the prompt comparison summarized in \S\ref{sec:abl}, where the main
text reports the CAR view (Fig.~\ref{fig:carresp}).

\paragraph{Output-collapse diagnostic.}
To test whether reduced personalization merely reflects malformed or
collapsed recommendations, we examine list length, within-response title
repetition, corpus-level diversity, and overlap with the unablated
response. At exact projection ($\alpha=1$), responses retain five-item
lists with within-response distinct-title ratios of $0.999$--$1.000$;
their corpus-level diversity is at least as high as under the prompt in
every model. They overlap with the unablated recommendations by
$0.76$--$0.89$, at least as much as under the prompt baseline
($0.69$--$0.84$). At $\alpha=5$, overlap falls to $0.39$--$0.59$,
indicating that amplified
removal causes broader output changes. These diagnostics rule out gross
formatting or repetition collapse at exact projection but do not
constitute a user-level evaluation of recommendation quality.
\section{Selectivity of the Direction Ablation}
\label{sec:app-orth}

\S\ref{sec:abl} reports whether removing one dimension's direction spares
the co-present dimension. This appendix records why that question can only
be answered on the taxonomy metric, and how the answer varies by model.

\paragraph{Semantic distance cannot express selectivity.}
For an $\mathrm{ablate}_A$ intervention, the natural semantic selectivity
measure is $\Delta_{\mathrm{pure}A} - \Delta_{\mathrm{pure}B}$, the shift
away from the $A$-target minus the shift away from the $B$-target. This
quantity is uninformative by construction: any edit that changes the
response at all moves it away from both pure references, so the two shifts
rise together whatever the edit removed. Empirically they do, at every
$\alpha$ we sweep, including exact projection. Semantic distance therefore
registers that the intervention acted, not which dimension it acted on.
CAR counts content per dimension and can separate the two.

\paragraph{Confinement holds on two models of three.}
On Mistral-7B, ablating the gender direction lowers gender-tagged content
by $0.169$ on race$\times$gender and $0.039$ on gender$\times$age while
race- and age-tagged content moves by $+0.004$ and $-0.003$; across the
full sweep of both cells the partner dimension's tagged rate never falls at
any $\alpha$. Llama-3-8B shows the same in the other direction: ablating
race lowers race-tagged content by $0.030$ with gender-tagged content at
$+0.002$, again never falling across the sweep. Neither operating point was
chosen for selectivity---each is the $\alpha$ that maximizes the target
reduction---so these are unselected demonstrations that the removal can be
confined to one dimension.

Qwen3-8B is the exception. At its taxonomy-optimal strengths the two
dimensions fall together: ablating race on race$\times$age lowers
race-tagged content by $0.130$ and age-tagged content by $0.104$, and on
gender$\times$age the off-target move ($-0.128$, $p=2.9\times10^{-15}$)
exceeds the target one ($-0.030$). On this model the intervention is better
described as suppressing personalization along a shared identity axis than
as a dimension-specific edit. Target and partner CAR values for all three
models are released with the code.

\paragraph{Why confinement is partial.}
That the removal is dimension-specific on some models and not others is
consistent with the composition result of \S\ref{sec:lin}. The mixed-cue
direction is substantially but not fully a linear combination of the
single-cue directions ($\bar\rho \in [0.576, 0.704]$), so the per-dimension
directions are not mutually orthogonal, and a rank-one projection along one
of them removes some of the other. We read this as cue-associated
directions inhabiting overlapping subspaces rather than clean orthogonal
axes, consistent with the linear-representation and superposition
literature \citep{arditi2024refusal, marshall2024refusal}. Making the
confinement hold across models remains open.

\end{document}